\batchmode
\documentclass[english]{IEEEtran}
\usepackage[T1]{fontenc}
\usepackage[latin9]{inputenc}
\usepackage{array}
\usepackage{multirow}
\usepackage{amsmath}
\usepackage{amsthm}
\usepackage{amssymb}
\usepackage{graphicx}

\makeatletter

\providecommand{\tabularnewline}{\\}

\theoremstyle{plain}
\newtheorem{thm}{\protect\theoremname}
\theoremstyle{definition}
\newtheorem{defn}[thm]{\protect\definitionname}
\theoremstyle{plain}
\newtheorem{prop}[thm]{\protect\propositionname}
\theoremstyle{plain}
\newtheorem{lem}[thm]{\protect\lemmaname}

\usepackage{cite}
\allowdisplaybreaks 

\usepackage[margin=8pt,font=footnotesize]{caption}
\usepackage{algorithm}
\usepackage{algpseudocode}
\usepackage{amsmath}  

\usepackage{tikz}
\usetikzlibrary{shapes,decorations}
\usepackage{subcaption}

\makeatother

\usepackage{babel}
\providecommand{\definitionname}{Definition}
\providecommand{\lemmaname}{Lemma}
\providecommand{\propositionname}{Proposition}
\providecommand{\theoremname}{Theorem}

\begin{document}
\title{A time-weighted metric for sets of trajectories to assess multi-object
tracking algorithms}
\author{Ángel F. García-Fernández$^{\star}$, Abu Sajana Rahmathullah$^{\circ}$,
Lennart Svensson$^{\dagger}$ \\
{\normalsize{}$^{\star}$Dept. of Electrical Engineering and Electronics,
University of Liverpool, United Kingdom}\\
{\normalsize{}$^{\star}$ARIES Research Center, Universidad Antonio
de Nebrija, Spain}\\
{\normalsize{}$^{\circ}$Zenseact, Sweden}\\
$^{\dagger}${\normalsize{}Dept. of Electrical Engineering, Chalmers
University of Technology, Sweden}\\
{\normalsize{}Emails: angel.garcia-fernandez@liverpool.ac.uk, abusajana@gmail.com,
lennart.svensson@chalmers.se }}

\maketitle
\thispagestyle{empty}
\begin{abstract}
This paper proposes a metric for sets of trajectories to evaluate
multi-object tracking algorithms that includes time-weighted costs
for localisation errors of properly detected targets, for false targets,
missed targets and track switches. The proposed metric extends the
metric in \cite{Angel20_d} by including weights to the costs associated
to different time steps. The time-weighted costs increase the flexibility
of the metric \cite{Angel20_d} to fit more applications and user
preferences. We first introduce a metric based on multi-dimensional
assignments, and then its linear programming relaxation, which is
computable in polynomial time and is also a metric. The metrics can
also be extended to metrics on random finite sets of trajectories
to evaluate and rank algorithms across different scenarios, each with
a ground truth set of trajectories.
\end{abstract}


\section{Introduction}

Multiple object tracking (MOT) algorithms play a fundamental role
in many applications, such as self-driving vehicles, robotics and
surveillance \cite{Mahler_book14}. In MOT, there is an unknown and
time-varying number of objects that appear in the surveillance area,
move following a certain trajectory and disappear from the surveillance
area. This ground truth can therefore be represented as a set of trajectories
\cite{Angel20_b}. The objective of MOT algorithms is to estimate
this set of trajectories based on noisy sensor data. An important
task is then to rank the performances of different algorithms by measuring
how similar the estimated set of trajectories is to the ground truth
\cite{Milan16_arxiv}. 

The quantification of the error between the ground truth set of trajectories
and the estimate is usually based on a distance function. For principled
evaluation and an intuitive notion of error, it is desirable that
this distance function is a metric in a mathematical sense\footnote{While distance and metric usually refer to the same concept, in this
paper, a distance is a function that does not necessarily meet the
metric properties.} \cite{Apostol_book74}. In addition, we are interested in metrics
that penalise the following important aspects of MOT: localisation
errors for properly detected targets, the number of missed and false
targets, and track switches \cite{Blackman_book99,Fridling91,Drummond92}.

For the related problem of multi-object filtering, in which we aim
to estimate the set of targets at the current time step, the optimal
subpattern assignment (OSPA) metric is widely used \cite{Schuhmacher08_b,Schuhmacher08}.
However, this metric does not penalise missed and false targets, e.g.,
it can be insensitive to the addition of false targets, and it can
promote algorithms with spooky effect at a distance \cite{Angel19_d}.
The generalised OSPA (GOSPA) metric (with parameter $\alpha=2$) penalises
localisation errors for properly detected targets, and the number
of missed and false targets, as typically required in MOT performance
evaluation \cite{Rahmathullah17}. In the GOSPA metric, we look for
an optimal assignment between the targets in the ground truth and
estimated targets, leaving missed and false targets unassigned.

For sets of trajectories, apart from penalising the errors for the
estimated set of targets at each time step, we are usually interested
in penalising track switches \cite[Sec. 13.6]{Blackman_book99}. That
is, the optimal assignments between estimated trajectories and real
trajectories may change at different time steps giving rise to undesired
track switches. There are also a number of distance functions for
sets of trajectories that are not metrics in the literature, for example,
\cite{Bernardin08,Ristani16,Luiten20,Ristic11,Canavan09,Silbert09,Manson92}. 

The OSPA$^{(2)}$ metric is a OSPA metric with a specific base distance
applied to sets of trajectories \cite{Beard20}. The OSPA$^{(2)}$
metric establishes an assignment between real and estimated trajectories
that is not allowed to change with time. This approach avoids track
switching considerations and may not rank different estimated sets
as desired in standard MOT performance evaluation, see Figure \ref{fig:OSPA-track-switching}.
As other metrics, OSPA$^{(2)}$ can be applied by subsampling the
set of trajectories at time steps of relevance \cite[Sec. V.B.2]{Beard20}.
In addition, for a single time step, OSPA$^{(2)}$ becomes the OSPA
metric for targets, which has the characteristics explained above.
\tikzset{xnodes/.style={cross out, scale=1.3}} 
\tikzset{ynodes/.style={circle, fill}} 
\tikzset{x=0.8cm,y=0.8cm, every text node part/.style={align=center}, every node/.style={font=\scriptsize, inner sep=1pt,outer sep=0pt, minimum size=2pt,  draw}}
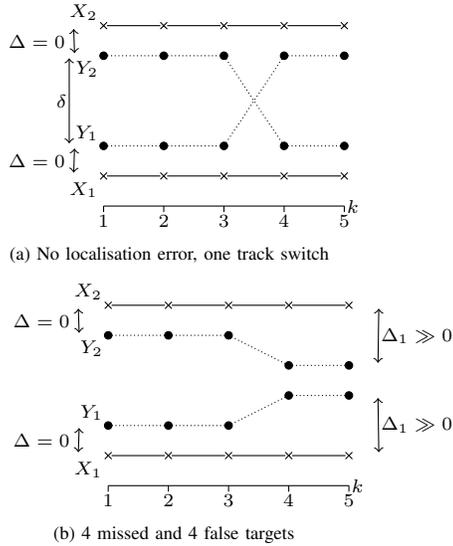
\begin{figure}	
\begin{center}
\begin{minipage}[t]{0.49\linewidth}
\centering
\begin{tikzpicture}
	\def \lenx {5}; 	\def \leny {5};\def \Del{0.5};\def \ylim {-0.5};
	\def \del {2};
	\node [xnodes, label = below left:$X_{1}$](x11) at (1, 0){};
	\foreach \x in {2,..., \lenx} {
		\node [xnodes](x1\x) at (\x, 0){};}
	\foreach \x in {2,..., \lenx} {
		\pgfmathtruncatemacro{\cur}{\x}
		\pgfmathtruncatemacro{\next}{\x - 1}
		\draw (x1\cur)--(x1\next);
		}
	
	\node [xnodes, label = above left:$X_{2}$](x21) at (1, \Del+\del){};
    \foreach \x in {2,..., \lenx} {
		\node [xnodes](x2\x) at (\x, \Del+\del){};}
	\foreach \x in {2,..., \lenx} {
		\pgfmathtruncatemacro{\cur}{\x}
		\pgfmathtruncatemacro{\next}{\x - 1}
		\draw (x2\cur)--(x2\next);
		}
	
	\node [ynodes, label = above left:$Y_{1}$](y11) at (1, \Del){};
	\foreach \x in {2,..., \leny} {
		\node [ynodes](y1\x) at (\x, \Del){};}
		
	\node [ynodes, label = below left:$Y_{2}$](y21) at (1, \del){};
	\foreach \x in {2,..., \leny} {
		\node [ynodes](y2\x) at (\x, \del){};}
	
    \draw [densely dotted](y11)--(y12)--(y13)--(y24)--(y25);
    \draw [densely dotted](y21)--(y22)--(y23)--(y14)--(y15);

	\draw [<->]([xshift=-10pt]x11.north west) -- ([xshift=-10pt]y11.south west) node[draw=none,fill=none,midway,left] {$\Delta=0\ $};
    \draw [<->]([xshift=-10pt]y21.north west) -- ([xshift=-10pt]x21.south west) node[draw=none,fill=none,midway,left] {$\Delta=0\ $};

	\draw [<->]([xshift=-12pt]y11.north west) -- ([xshift=-12pt]y21.south west) node[draw=none,fill=none,midway,left] {$\delta$};  
	
	\foreach \x in {1, ..., \lenx}{
	    \draw (\x , \ylim) -- (\x, \ylim-0.1) node[draw = none, below] {$\x$};
    }            
   \draw [-](1, \ylim) -- (\lenx, \ylim) node[draw=none, right] {$k$};
\end{tikzpicture}
\subcaption{No localisation error, one track switch}
\label{fig:1_ega}
\end{minipage}
\vspace{0.2cm}

\begin{minipage}[t]{0.49\linewidth}
\centering
\begin{tikzpicture}
	\def \lenx {5}; 	\def \leny {5};\def \Del{0.5};\def \ylim {-0.5};
	\def \del {2};
	\node [xnodes, label = below left:$X_{1}$](x11) at (1, 0){};
	\foreach \x in {2,..., \lenx} {
		\node [xnodes](x1\x) at (\x, 0){};}
	\foreach \x in {2,..., \lenx} {
		\pgfmathtruncatemacro{\cur}{\x}
		\pgfmathtruncatemacro{\next}{\x - 1}
		\draw (x1\cur)--(x1\next);
		}
	
	\node [xnodes, label = above left:$X_{2}$](x21) at (1, \Del+\del){};
    \foreach \x in {2,..., \lenx} {
		\node [xnodes](x2\x) at (\x, \Del+\del){};}
	\foreach \x in {2,..., \lenx} {
		\pgfmathtruncatemacro{\cur}{\x}
		\pgfmathtruncatemacro{\next}{\x - 1}
		\draw (x2\cur)--(x2\next);
		}
	
	\def \lenybreak {3}; 	\def \lenybreakl {4}; \def \Dell{1};
	\node [ynodes, label = above left:$Y_{1}$](y11) at (1, \Del){};
	\foreach \x in {2,..., \lenybreak} {
		\node [ynodes](y1\x) at (\x, \Del){};}		
		
	\foreach \x in {\lenybreakl,..., \leny} {
		\node [ynodes](y1\x) at (\x, \Dell){};}

	\foreach \x in {2,..., \leny} {
		\pgfmathtruncatemacro{\cur}{\x}
		\pgfmathtruncatemacro{\next}{\x - 1}
		\draw [densely dotted](y1\cur)--(y1\next);
		}
		
	\node [ynodes, label = below left:$Y_{2}$](y21) at (1, \del){};
	\foreach \x in {2,..., \lenybreak} {
		\node [ynodes](y2\x) at (\x, \del){};}
		
	\foreach \x in {\lenybreakl,..., \leny} {
		\node [ynodes](y2\x) at (\x, \del - \Del){};}

	\foreach \x in {2,..., \leny} {
		\pgfmathtruncatemacro{\cur}{\x}
		\pgfmathtruncatemacro{\next}{\x - 1}
		\draw [densely dotted](y2\cur)--(y2\next);
		}
   
   \draw [<->]([xshift=-10pt]x11.north west) -- ([xshift=-10pt]y11.south west) node[draw=none,fill=none,midway,left] {$\Delta=0\ $};
   \draw [<->]([xshift=-10pt]y21.north west) -- ([xshift=-10pt]x21.south west) node[draw=none,fill=none,midway,left] {$\Delta=0\ $};
   
	\draw [<->]([xshift=10pt]x15.north east) -- ([xshift=10pt]y15.south east) node[draw=none,fill=none,midway,right] {$\Delta_1 \gg 0\ $};
    \draw [<->]([xshift=10pt]y25.north east) -- ([xshift=10pt]x25.south east) node[draw=none,fill=none,midway,right] {$\Delta_1 \gg 0\ $};
	
	\foreach \x in {1, ..., \lenx}{
	    \draw (\x , \ylim) -- (\x, \ylim-0.1) node[draw = none, below] {$\x$};
    }            
   \draw [-](1, \ylim) -- (\lenx, \ylim) node[draw=none, right] {$k$};
\end{tikzpicture}
\subcaption{4 missed and 4 false targets}
\label{fig:1_egb}
\end{minipage}
\caption{Estimated set of trajectories $\mathbf{Y}=\{Y_1,Y_2\}$ against the ground truth $\mathbf{X}=\{X_1,X_2\}$. In (a),  $\mathbf{Y}$ has zero localisation errors and a track switch. In (b), $\mathbf{Y}$ has zero localisation errors up to time step 3, but there are 4 missed and false targets at time steps 4 and 5 (Distance $\Delta_1$ may be arbitrarily large). OSPA$^{(2)}$ penalises these two cases equally. With a low switching cost, according to the trajectory metrics \cite{Angel20_d}, (a) has a lower error than (b).}
\label{fig:OSPA-track-switching}
\end{center}
\vspace{-0.5cm}
\end{figure}

Bento's metrics \cite{Bento_draft16} for sets of trajectories penalise
track switches by first adding $\ast$-trajectories to the ground
truth and the estimate so that both sets have equal cardinality. The
track switches are penalised based on a track switching cost function
that penalises changes in the permutations that associate the (augmented)
estimated trajectories to the (augmented) true trajectories. A metric
that extends the GOSPA metric to sets of trajectories and penalises
the number of track switches based on changes in assignments/unassignments,
without using permutations and $\ast$-trajectories, was proposed
in \cite{Angel20_d}. The linear programming (LP) relaxation of this
metric is also a metric and is computable in polynomial time \cite{Angel20_d}.
We refer to these two metrics as trajectory metrics. These trajectory
metrics penalise localisation errors for properly detected targets,
and the number of missed, false targets, and track switches. 

In this paper, we increase the flexibility of the trajectory metrics
in \cite{Angel20_d} by adding time-weighted costs at each time step.
This can be useful, for instance, in online tracking where we generally
want to place more weight on estimates corresponding to recent time
steps \cite{Silbert09}. Time-weighted costs can also be useful to
measure the accuracy of a prediction of the set of trajectories at
future time steps, by placing more weight to closer time steps, similarly
to the use of discounted costs in sensor management \cite{Hero_book08}.
As in \cite{Angel20_d}, the time-weighted trajectory metrics can
also be extended to metrics on random finite sets (RFSs) of trajectories
\cite{Angel20_b}.

\section{Problem formulation and background\label{sec:Problem-formulation}}

In the standard multi-object tracking model, there is an unknown number
of objects at each time step in the surveillance area. Each object
is born at a certain time step, moves following a certain sequence
of states and dies \cite{Mahler_book14}. An object trajectory is
represented by a variable $X=\left(\omega,x^{1:\nu}\right)$ where
$\omega$ is its initial time step, $\nu$ is its length and $x^{1:\nu}=\left(x^{1},...,x^{\nu}\right)$
is its sequence of consecutive states \cite{Angel20_b}. We consider
trajectories from time step 1 to $T$, which implies that the variable
$\left(\omega,\nu\right)$ belongs to the set $I_{(T)}=\left\{ \left(\omega,\nu\right):1\leq\omega\leq T\,\mathrm{and}\,1\leq\nu\leq T-\omega+1\right\} $,
and variable $X\in\mathbb{T}_{\left(T\right)}$ where $\mathbb{T}_{\left(T\right)}=\uplus_{\left(\omega,\nu\right)\in I_{(T)}}\left\{ \omega\right\} \times\mathbb{R}^{\nu n_{x}}$
and $\uplus$ stands for the union of sets that are mutually disjoint\footnote{We consider trajectories from time step 1 to $T$, but it is trivial
to consider trajectories in a different time interval.}.

The variable of interest in MOT is the (ground truth) set of trajectories
up to time step $T$, which is denoted by $\mathbf{X}=\left\{ X_{1},...,X_{n_{\mathbf{X}}}\right\} $.
The set $\mathbf{X}$ belongs to $\Upsilon=\mathcal{F}\left(\mathbb{T}_{\left(T\right)}\right)$,
which denotes the set of all finite subsets of $\mathbb{T}_{\left(T\right)}$. 

Based on noisy sensor data, an MOT algorithm provides an estimated
set of trajectories $\mathbf{Y}=\left\{ Y_{1},...,Y_{n_{\mathbf{Y}}}\right\} $.
We would like to measure the distance between $\mathbf{Y}$ and $\mathbf{X}$
to evaluate estimation error. A mathematically principled way of measuring
distances is via metrics. 

A metric on $\Upsilon$ is a function $d\left(\cdot,\cdot\right):\Upsilon\times\Upsilon\rightarrow\left[0,\infty\right)$
such that \cite{Apostol_book74}
\begin{itemize}
\item $d\left(\mathbf{X},\mathbf{Y}\right)=0$ if and only if $\mathbf{X}=\mathbf{Y}$
(identity),
\item $d\left(\mathbf{X},\mathbf{Y}\right)=d\left(\mathbf{Y},\mathbf{X}\right)$
(symmetry),
\item $d\left(\mathbf{X},\mathbf{Y}\right)\leq d\left(\mathbf{X},\mathbf{Z}\right)+d\left(\mathbf{Z},\mathbf{Y}\right)$
(triangle inequality).
\end{itemize}
The aim of this paper is to propose a metric on the space $\Upsilon$.
While the metric is designed for the space $\Upsilon$, which considers
trajectories without holes, the proposed metric can also be used with
trajectories with holes \cite[Sec. II.A]{Angel20_d}.

In most applications, the main aim of metrics is to rank different
algorithms (each with a different $\mathbf{Y}$) against a ground
truth set of trajectories (a single $\mathbf{X}$). The case in which
we have a data set with several scenarios (each with a different ground
truth) can be addressed via metrics on RFSs of trajectories, and will
be explained in Section \ref{sec:Metrics-on-RFS}.

\subsection{GOSPA metric}

We review the GOSPA metric as it is a building block for the metric
for sets of trajectories. Let us consider the sets of targets $\ensuremath{\mathbf{x}=\left\{ x_{1},...,x_{n_{\mathbf{x}}}\right\} }$
and $\ensuremath{\mathbf{y}=\left\{ y_{1},...,y_{n_{\mathbf{y}}}\right\} }$.
We denote the set of all possible assignment sets between sets $\left\{ 1,..,n_{\mathbf{x}}\right\} $
and $\left\{ 1,...,n_{\mathbf{y}}\right\} $ as $\Gamma_{\mathbf{x},\mathbf{y}}$.
That is, any set $\theta\in\Gamma_{\mathbf{x},\mathbf{y}}$ is such
that $\theta\subseteq\left\{ 1,..,n_{\mathbf{x}}\right\} \times\left\{ 1,..,n_{\mathbf{y}}\right\} $
and, $\left(i,j\right),\left(i,j'\right)\in\theta$ implies $j=j'$
and, $\left(i,j\right),\left(i',j\right)\in\theta$ implies $i=i'$. 
\begin{defn}
\label{def:GOSPA_alpha2}\textit{Given a base metric $d_{b}\left(\cdot,\cdot\right)$
on $\mathbb{R}^{n_{x}}$, a scalar $c>0$, and a scalar $p$ with
$1\leq p<\infty$, the GOSPA metric ($\alpha=2$) between sets of
targets $\mathbf{x}$ and $\mathbf{y}$ is} \cite[Prop. 1]{Rahmathullah17}
\begin{align}
 & d_{G}\left(\mathbf{x},\mathbf{y}\right)\nonumber \\
 & =\min_{\theta\in\Gamma_{\mathbf{x},\mathbf{y}}}\left(\sum_{\left(i,j\right)\in\theta}d_{b}\left(x_{i},y_{j}\right)^{p}+\frac{c^{p}}{2}\left(\left|\mathbf{x}\right|+\left|\mathbf{y}\right|-2\left|\theta\right|\right)\right)^{1/p}.\label{eq:GOSPA_alpha2}
\end{align}
\textit{where $\left|\mathbf{x}\right|$ denotes the number of elements
in the set $\mathbf{x}$.}
\end{defn}
The first term in (\ref{eq:GOSPA_alpha2}) corresponds to the localisation
errors (to the $p$-th power) for assigned targets (properly detected
targets), which meet $\left(i,j\right)\in\theta$. The terms $\frac{c^{p}}{2}\left(\left|\mathbf{x}\right|-\left|\theta\right|\right)$
and $\frac{c^{p}}{2}\left(\left|\mathbf{y}\right|-\left|\theta\right|\right)$
are the costs (to the $p$-th power) for missed and false targets,
respectively. We have also dropped the dependence of $d_{G}\left(\cdot,\cdot\right)$
on $p$ and $c$ for notational simplicity. 

In particular, the trajectory metrics use GOSPA with sets $\mathbf{x}$
and $\mathbf{y}$ with at most one element, such that $\left|\mathbf{x}\right|\leq1$
and $\left|\mathbf{y}\right|\leq1$. In this case, (\ref{eq:GOSPA_alpha2})
becomes
\begin{align}
d_{G}\left(\mathbf{x},\mathbf{y}\right) & \triangleq\begin{cases}
\min\left(c,d_{b}\left(x,y\right)\right) & \mathbf{x}=\left\{ x\right\} ,\mathbf{y}=\left\{ y\right\} \\
0 & \mathbf{x}=\mathbf{y}=\emptyset\\
\frac{c}{2^{1/p}} & \mathrm{otherwise.}
\end{cases}\label{eq:baseMetric}
\end{align}

\section{Time-weighted trajectory metrics\label{sec:Time-weighted-trajectory-metrics}}

This section presents the time-weighted trajectory metrics. Preliminary
notation is introduced in Section \ref{subsec:Notation}. The time-weighted
multi-dimensional assignment and linear programming relaxation metrics
are proposed in Sections \ref{subsec:Time-weighted-multi-dimensional_metric}
and \ref{subsec:Time-weighted-LP-metric}, respectively. 

\subsection{Notation\label{subsec:Notation}}

We introduce the following notation to define the metrics. Given a
single trajectory $X=\left(\omega,x^{1:\nu}\right)$, the set with
the state of the trajectory at time step $k$ is
\begin{align}
\tau^{k}\left(X\right)\triangleq\begin{cases}
\{x^{k+1-\omega}\} & \omega\leq k\leq\omega+\nu-1\\
\emptyset & \text{otherwise}.
\end{cases}
\end{align}
Similarly, given a set $\mathbf{X}$ of trajectories, the set $\tau^{k}\left(\mathbf{X}\right)$
of target states at time $k$ is
\begin{equation}
\tau^{k}\left(\mathbf{X}\right)=\bigcup_{X\in\mathbf{X}}\tau^{k}\left(X\right).\label{eq:set_targets_formula}
\end{equation}
The sets of targets corresponding to trajectories $X_{i}$ and $Y_{j}$
at time step $k$ are denoted by $\mathbf{x}_{i}^{k}=\tau^{k}\left(X_{i}\right)$
and $\mathbf{y}_{j}^{k}=\tau^{k}\left(Y_{j}\right)$, respectively.
These sets are such that $\left|\mathbf{x}_{i}^{k}\right|\leq1$ and
$\left|\mathbf{y}_{i}^{k}\right|\leq1$.

In the trajectory metrics, there are assignments at a trajectory level
to determine track switches. That is, at each time step, each target
set $\mathbf{x}_{i}^{k}$ is assigned to a target set $\mathbf{y}_{i}^{k}$
or is left unassigned. The set of all possible assignment vectors
between sets $\{1,\ldots,n_{\mathbf{X}}\}$ and $\ensuremath{\{0,\ldots,n_{\mathbf{Y}}\}}$
is denoted by $\ensuremath{\Pi_{\mathbf{X},\mathbf{Y}}}.$ The assignment
vector $\ensuremath{\pi^{k}=[\pi_{1}^{k},...,\pi_{n_{\mathbf{X}}}^{k}]\in\Pi_{\mathbf{X},\mathbf{Y}}}$
meets $\pi_{i}^{k}\in\{0,\ldots,n_{\mathbf{X}}\}$ with the constraint
that $\pi_{i}^{k}=\pi_{i^{\prime}}^{k}=j>0$ implies $i=i^{\prime}$.
Here, $\pi_{i}^{k}=j\neq0$ implies that $X_{i}$ is assigned to $Y_{j}$
at time $k$ and $\pi_{i}^{k}=0$ implies that $X_{i}$ is unassigned
at time $k$. This definition of assignment vectors implies that,
at a specific time step, there can only be one trajectory $Y_{j}$
assigned to a trajectory $X_{i}$, but multiple trajectories in $\mathbf{X}$
and $\mathbf{Y}$ can be left unassigned.

\subsection{Time-weighted multi-dimensional assignment metric\label{subsec:Time-weighted-multi-dimensional_metric}}

The proposed time-weighted metric based on multi-dimensional assignment
is given by the following definition.
\begin{defn}
\label{def:multi-dimensional-metric} \textit{For $1\leq p<\infty$,
cut-off parameter $c>0$, switching penalty $\gamma>0$, base metric
$d_{b}\left(\cdot,\cdot\right)$ in the single-target space, localisation
time weight $w_{1}^{k}>0$ at time step $k$ and switching time weight
$w_{2}^{k}>0$ at time step $k$, the time-weighted multi-dimensional
assignment metric $d\left(\mathbf{X},\mathbf{Y}\right)$ between two
sets $\mathbf{X}$ and $\mathbf{Y}$ of trajectories is}
\begin{align}
d\left(\mathbf{X},\mathbf{Y}\right) & =\min_{\substack{\pi^{k}\in\Pi_{\mathbf{X},\mathbf{Y}}\\
k=1,\ldots,T
}
}\left(\sum_{k=1}^{T}w_{1}^{k}d_{\mathbf{X},\mathbf{Y}}^{k}\left(\mathbf{X},\mathbf{Y},\pi^{k}\right)^{p}\right.\nonumber \\
 & \:\left.+\sum_{k=1}^{T-1}w_{2}^{k}s_{\mathbf{X},\mathbf{Y}}\left(\pi^{k},\pi^{k+1}\right)^{p}\right)^{1/p}\label{eq:multi-dimensional-metric}
\end{align}
\textit{where the costs (to the $p$-th power) for properly detected
targets, missed targets and false targets at time step $k$ are}
\begin{align}
d_{\mathbf{X},\mathbf{Y}}^{k}\left(\mathbf{X},\mathbf{Y},\pi^{k}\right)^{p} & =\sum_{\left(i,j\right)\in\theta^{k}\left(\pi^{k}\right)}d_{G}\left(\mathbf{x}_{i}^{k},\mathbf{y}_{j}^{k}\right)^{p}\nonumber \\
 & +\frac{c^{p}}{2}\left(\left|\tau^{k}\left(\mathbf{X}\right)\right|+\left|\tau^{k}\left(\mathbf{Y}\right)\right|-2\left|\theta^{k}\left(\pi^{k}\right)\right|\right)\label{eq:GOSPA_no_minimisation}
\end{align}
\textit{with}
\begin{align}
\theta^{k}\left(\pi^{k}\right) & =\left\{ (i,\pi_{i}^{k}):i\in\{1,\ldots,n_{\mathbf{X}}\},\vphantom{d\left(\mathbf{x}_{i}^{k},\mathbf{y}_{\pi_{i}^{k}}^{k}\right)}\right.\nonumber \\
 & \,\left.\left|\mathbf{x}_{i}^{k}\right|=\left|\mathbf{y}_{\pi_{i}^{k}}^{k}\right|=1,\,d_{G}\left(\mathbf{x}_{i}^{k},\mathbf{y}_{\pi_{i}^{k}}^{k}\right)<c\right\} \label{eq:target_level_assignment}
\end{align}
\textit{and the switching cost (to the $p$-th power) from time step
$k$ to $k+1$ is}
\begin{equation}
s_{\mathbf{X},\mathbf{Y}}(\pi^{k},\pi^{k+1})^{p}=\gamma^{p}\sum_{i=1}^{n_{\mathbf{X}}}s\big(\pi_{i}^{k},\pi_{i}^{k+1}\big)\label{eq:switching_cost_MD}
\end{equation}
\begin{align*}
s\big(\pi_{i}^{k},\pi_{i}^{k+1}\big) & =\begin{cases}
0 & \pi_{i}^{k}=\pi_{i}^{k+1}\\
1 & \pi_{i}^{k}\neq\pi_{i}^{k+1},\pi_{i}^{k}\neq0,\pi_{i}^{k+1}\neq0\\
\frac{1}{2} & \text{otherwise. \quad\quad\quad\quad\quad\quad\quad\quad}
\end{cases}
\end{align*}
\end{defn}
In the metric (\ref{eq:multi-dimensional-metric}), we have an assignment
vector $\pi^{k}$ at a trajectory level at time step $k$. The change
from $\pi^{k}$ to $\pi^{k+1}$ determines the switching cost (to
the $p$-th power) in (\ref{eq:switching_cost_MD}). If $\pi_{i}^{k}=\pi_{i}^{k+1}$,
there is no switching cost for the $i$-th trajectory in the ground
truth. If $\pi_{i}^{k}\neq\pi_{i}^{k+1}$ and both $\pi_{i}^{k}$
and $\pi_{i}^{k+1}$ are different from zero, which implies $X_{j}$
is assigned to two different trajectories in $\mathbf{Y}$ at time
steps $k$ and $k+1$, the switching cost (to the $p$-th power) is
$\gamma^{p}$. Finally, if $\pi_{i}^{k}\neq\pi_{i}^{k+1}$ and either
$\pi_{i}^{k}$ or $\pi_{i}^{k+1}$ is zero, which implies that $X_{j}$
is unassigned at either time step $k$ or $k+1$, and assigned to
a trajectory in $\mathbf{Y}$ at the other time step, the switching
cost (to the $p$-th power) is $\gamma^{p}/2$. This situation represents
a half track switch, in which the switching cost from assigned to
unassigned must be half the switching cost from assigned to assigned.
This is a necessary property of a trajectory metric based on penalising
switches with assignment vectors \cite{Angel20_d}.

Equation (\ref{eq:GOSPA_no_minimisation}) corresponds to the GOSPA
metric in (\ref{eq:GOSPA_alpha2}) without the minimisation over the
target-level assignments. Instead, the GOSPA assignment (target-level
assignment) is fixed by the assignment vector at a trajectory level
$\pi^{k}$, excluding assignments with targets whose distance is greater
than $c$, see (\ref{eq:target_level_assignment}). Then, (\ref{eq:GOSPA_no_minimisation})
includes the localisation costs for properly detected targets (those
included in $\theta^{k}\left(\pi^{k}\right)$), and costs for missed
and false targets, each of them penalised with a cost $\frac{c^{p}}{2}$,
as in (\ref{eq:GOSPA_alpha2}). The reason for using both trajectory
and target level assignments is explained in \cite[Sec. III.B]{Angel20_d}.

The proof that (\ref{eq:multi-dimensional-metric}) is a metric is
analogous to the proof of the linear programming metric, which is
given by Proposition \ref{prop:LP_metric} in Section \ref{subsec:Time-weighted-LP-metric}.
Equation (\ref{eq:multi-dimensional-metric}) coincides with the multi-dimensional
trajectory metric in \cite{Angel20_d} by setting $w_{1}^{k}=1$ and
$w_{2}^{k}=1$ for all $k$. It is also possible to have a time-dependent
base metric and keep the metric properties.

\subsection{Time-weighted linear programming relaxation metric\label{subsec:Time-weighted-LP-metric}}

The assignment vectors in Definition \ref{def:multi-dimensional-metric}
can also be written in terms of binary weight matrices \cite{Angel20_d}.
We use $\mathcal{W}_{\mathbf{X},\mathbf{Y}}$ to denote the set of
all binary matrices of dimension $(n_{\mathbf{X}}+1)\times(n_{\mathbf{Y}}+1)$,
representing assignments between $\mathbf{X}$ and $\mathbf{Y}$.
A matrix $W^{k}\in\mathcal{W}_{\mathbf{X},\mathbf{Y}}$ satisfies
the following properties:
\begin{align}
\sum_{i=1}^{n_{\mathbf{X}}+1}W^{k}(i,j) & =1,\ j=1,\ldots,n_{\mathbf{Y}}\label{eq:binary_constraint1}\\
\sum_{j=1}^{n_{\mathbf{Y}}+1}W^{k}(i,j) & =1,\ i=1,\ldots,n_{\mathbf{X}}\\
W^{k}(n_{\mathbf{X}}+1,n_{\mathbf{Y}}+1) & =0,\label{eq:binary_constraint3}\\
W^{k}(i,j) & \in\{0,1\},\forall\ i,j\label{eq:binary_constraint4}
\end{align}
where $W^{k}(i,j)$ represents the element in the row $i$ and column
$j$ of matrix $W^{k}$. We have $W^{k}(i,j)=1$ if $\mathbf{x}_{i}^{k}$
is associated to $\mathbf{y}_{j}^{k}$, $W^{k}(i,n_{\mathbf{Y}}+1)=1$
if $\mathbf{x}_{i}^{k}$ is unassigned, and $W^{k}(n_{\mathbf{X}}+1,j)=1$
if $\mathbf{y}_{j}^{k}$ is unassigned. The bijection between the
assignment vectors and the binary weight matrices is given in \cite[Sec. IV.A]{Angel20_d},
and how to write Definition \ref{def:multi-dimensional-metric} in
terms of these matrices is direct from Lemma 1 in \cite{Angel20_d}.

The binary constraint (\ref{eq:binary_constraint4}) can be relaxed
to define the set $\mathcal{\overline{W}}_{\mathbf{X},\mathbf{Y}}$
of matrices. A matrix $W^{k}\in\mathcal{\overline{W}}_{\mathbf{X},\mathbf{Y}}$
meets (\ref{eq:binary_constraint1})-(\ref{eq:binary_constraint3})
and
\begin{align}
W^{k}(i,j)\geq0, & \quad\forall i,j.
\end{align}

\begin{prop}
\label{prop:LP_metric}For $1\leq p<\infty$, cut-off $c>0$, switching
penalty $\gamma>0$, base metric $d_{b}\left(\cdot,\cdot\right)$
on the single-target space, localisation time weight $w_{1}^{k}>0$
at time step $k$ and switching time weight $w_{2}^{k}>0$ at time
step $k$, the LP relaxation $\overline{d}\left(\mathbf{X},\mathbf{Y}\right)$
of $d\left(\mathbf{X},\mathbf{Y}\right)$ is also a metric and is
given by
\begin{align}
\overline{d}\left(\mathbf{X},\mathbf{Y}\right) & =\min_{\substack{W^{k}\in\mathcal{\overline{W}}_{\mathbf{X},\mathbf{Y}}\\
k=1,\ldots,T
}
}\Bigg(\sum_{k=1}^{T}w_{1}^{k}\mathrm{tr}\big[\big(D_{\mathbf{X},\mathbf{Y}}^{k}\big)^{\dagger}W^{k}\big]\nonumber \\
 & \quad+\frac{\gamma^{p}}{2}\sum_{k=1}^{T-1}w_{2}^{k}\sum_{i=1}^{n_{\mathbf{X}}}\sum_{j=1}^{n_{\mathbf{Y}}}|W^{k}(i,j)-W^{k+1}(i,j)|\Bigg)^{\frac{1}{p}},\label{eq:LP_metric}
\end{align}
where $D_{\mathbf{X},\mathbf{Y}}^{k}$ is a $(n_{\mathbf{X}}+1)\times(n_{\mathbf{Y}}+1)$
matrix whose $(i,j)$ element is
\begin{align}
D_{\mathbf{X},\mathbf{Y}}^{k}(i,j) & =d_{G}\left(\mathbf{x}_{i}^{k},\mathbf{y}_{j}^{k}\right)^{p}.
\end{align}
\end{prop}
The proof of Proposition \ref{prop:LP_metric} is given in Appendix
\ref{sec:AppendixA}. It should be noted that if we use $\mathcal{W}_{\mathbf{X},\mathbf{Y}}$
instead of $\mathcal{\overline{W}}_{\mathbf{X},\mathbf{Y}}$ in (\ref{eq:LP_metric}),
we recover the metric (\ref{eq:multi-dimensional-metric}). How to
write (\ref{eq:LP_metric}) as an LP is direct from \cite[App. B.B]{Angel20_d}.
We can also use clustering to compute the time-weighted metrics \cite[Sec. IV.D]{Angel20_d}. 

\section{Metric properties\label{sec:Metric-properties}}

This section explains several properties of the metrics. We discuss
possible choices of weights for different applications in Section
\ref{subsec:Choice-of-weights}. The decomposition of the metrics
into the different costs is explained in Section \ref{subsec:Metric-decomposition}.
We provide some inequalities for the metrics in Section \ref{subsec:Inequalities}.

\subsection{Choice of weights\label{subsec:Choice-of-weights}}

We proceed to discuss several options to choose the weights that may
be useful in some applications.

\subsubsection{Online trackers\label{subsec:Online-trackers}}

In online MOT algorithms, at each time step, we have an estimate of
the set of trajectories, and a ground truth set of trajectories. These
sets of trajectories may include all trajectories up to the current
time or only the ones that are present at the current time step \cite{Granstrom18}.
In some applications, we may want to place more emphasis on recent
time steps to evaluate the performance of different trackers, as illustrated
in Figure \ref{fig:Illustration_time_weighted_metric}. We can achieve
this using the time-weighted trajectory metrics with an exponential
weight
\begin{equation}
w_{1}^{k}=\rho^{T-k}\label{eq:weight1_online_tracker}
\end{equation}
where $\rho\in\left(0,1\right)$ is a forgetting factor, and $w_{2}^{k}=w_{1}^{k+1}$
for $k\in\left\{ 1,...,T-1\right\} $. Then, the different costs at
the current time step $T$ have weight 1 and, as we move backwards
in time, the different costs are dampened with an exponential factor
$\rho$. The lower $\rho$ is, the more we disregard the accuracy
of the estimation of the trajectories in the past to determine the
error.

\tikzset{xnodes/.style={cross out, scale=1.3}} 
\tikzset{ynodes/.style={circle, fill}} 
\tikzset{x=0.8cm,y=0.8cm, every text node part/.style={align=center}, every node/.style={font=\scriptsize, inner sep=1pt,outer sep=0pt, minimum size=2pt,  draw}}
\begin{figure}[th!]
\begin{center}
\begin{minipage}[t]{0.48\textwidth}
\centering
\begin{tikzpicture}
	\def \lenx {6}; 	\def \leny {5};\def \Del{0.5};\def \ylim {-0.5};
	\def \del {2};
	\node [xnodes, label = below left:$X$](x11) at (1, 0){};
	\foreach \x in {2,..., \lenx} {
		\node [xnodes](x1\x) at (\x, 0){};}
	\foreach \x in {2,..., \lenx} {
		\pgfmathtruncatemacro{\cur}{\x}
		\pgfmathtruncatemacro{\next}{\x - 1}
		\draw (x1\cur)--(x1\next);
		}
		
	\node [ynodes, label = above left:$Y$](y11) at (1, \Del){};
	\foreach \x in {2,..., \leny} {
		\node [ynodes](y1\x) at (\x, \Del){};}
	\foreach \x in {2,..., \leny} {
		\pgfmathtruncatemacro{\cur}{\x}
		\pgfmathtruncatemacro{\next}{\x - 1}
		\draw [densely dotted] (y1\cur)--(y1\next);
		}			

	\draw [<->]([xshift=-10pt]x11.north west) -- ([xshift=-10pt]y11.south west) node[draw=none,fill=none,midway,left] {$\Delta\ $};
	
	\foreach \x in {1, ..., \lenx}{
	    \draw (\x , \ylim) -- (\x, \ylim-0.1) node[draw = none, below] {$\x$};
    }            
   \draw [-](1, \ylim) -- (\lenx, \ylim) node[draw=none, right] {$k$};
\end{tikzpicture}
\subcaption{One missed target at time step 6.}
\label{fig:3_ega}
\end{minipage}
\vspace{0.5cm}

\begin{minipage}[t]{0.48\textwidth}
\centering
\begin{tikzpicture}
	\def \lenx {6}; 	\def \leny {5};\def \Del{0.5};\def \ylim {-0.5};
	\def \del {2};
	\node [xnodes, label = below left:$X$](x11) at (1, 0){};
	\foreach \x in {2,..., \lenx} {
		\node [xnodes](x1\x) at (\x, 0){};}
	\foreach \x in {2,..., \lenx} {
		\pgfmathtruncatemacro{\cur}{\x}
		\pgfmathtruncatemacro{\next}{\x - 1}
		\draw (x1\cur)--(x1\next);
		}
		
	\def \shift {1};
	\node [ynodes, label = above left:$Y$](y11) at (1 + \shift, \Del){};
	\foreach \x in {2,..., \leny} {
		\node [ynodes](y1\x) at (\x + \shift, \Del){};}
	\foreach \x in {2,..., \leny} {
		\pgfmathtruncatemacro{\cur}{\x}
		\pgfmathtruncatemacro{\next}{\x - 1}
		\draw [densely dotted] (y1\cur)--(y1\next);
		}			
		
	\draw [<->]([xshift=10pt]x16.north west) -- ([xshift=10pt]y15.south west) node[draw=none,fill=none,midway,right] {$\Delta\ $};	
				
	\foreach \x in {1, ..., \lenx}{
	    \draw (\x , \ylim) -- (\x, \ylim-0.1) node[draw = none, below] {$\x$};
    }            
   \draw [-](1, \ylim) -- (\lenx, \ylim) node[draw=none, right] {$k$};
\end{tikzpicture}
\subcaption{One missed target at time step 1}
\label{fig:3_egb}
\end{minipage}
\caption{Two estimated sets of trajectories against a ground truth (both sets with a single trajectory), with $\Delta \ll c$. In some applications, it may be suitable to penalise missed targets at different time instants differently. If we use the metric to compare the estimated sets of trajectories of different trackers at the current time step, a missed target in the past may be less important than a missed target at the current time step. In this setting, (b) can be considered a better estimate than (a).}
\label{fig:Illustration_time_weighted_metric}
\end{center}
\vspace{-0.5cm}
\end{figure}
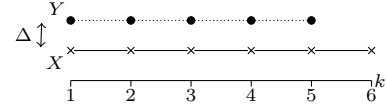
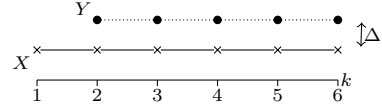

We may also want to normalise the weight (\ref{eq:weight1_online_tracker})
to sum to one in the considered window such that
\begin{equation}
w_{1}^{k}=\frac{\rho^{T-k}}{\sum_{j=1}^{T}\rho^{T-j}}=\frac{1-\rho}{1-\rho^{T}}\rho^{T-k}\label{eq:weight1_online_tracker_norm}
\end{equation}
and $w_{2}^{k}=w_{1}^{k+1}$ for $k\in\left\{ 1,...,T-1\right\} $. 

\subsubsection{Predictors}

In some applications, we are interested in predicting the set of trajectories
in a future time window \cite{Granstrom19_prov2,Tang19b}, for example,
in collision avoidance systems. For measuring the error of different
predictors, we can choose a time window starting at the following
time step and a weight 
\begin{equation}
w_{1}^{k}=\rho^{k-1}\label{eq:weight1_predictor}
\end{equation}
where $\rho\in\left(0,1\right)$, $w_{2}^{k}=w_{1}^{k+1}$ for $k\in\left\{ 1,...,T-1\right\} $,
and index $k$ goes through the future time steps. In this case, costs
at the next time step have weight 1 and, as we move forward in time,
the costs decrease with an exponential factor $\rho$. 

As in Section \ref{subsec:Online-trackers}, we may also want to normalise
the weight (\ref{eq:weight1_predictor}) such that
\begin{equation}
w_{1}^{k}=\frac{\rho^{k-1}}{\sum_{j=1}^{T}\rho^{k-1}}=\frac{1-\rho}{1-\rho^{T}}\rho^{k-1}
\end{equation}
and $w_{2}^{k}=w_{1}^{k+1}$ for $k\in\left\{ 1,...,T-1\right\} $.

\subsubsection{Time-dependent sampling intervals}

There are applications in which the time interval between measurements
is not constant. Let $t_{k}$ be the sampling time corresponding to
the $k$-th measurement such that the time interval between consecutive
measurements is $\Delta t_{k}=t_{k}-t_{k-1}$, with $t_{0}=0$. In
this setting, the (sampled) set of trajectories is represented as
in Section \ref{sec:Problem-formulation}, and each trajectory contains
state information at the sampled time steps \cite{Angel20_f}.

In this scenario, it may be useful to set weights equal or proportional
to the time interval between measurements, for example, $w_{1}^{k}=\Delta t_{k}$
and $w_{2}^{k}=w_{1}^{k+1}$ for $k\in\left\{ 1,...,T-1\right\} $.
With this approach, we give more weight to the time steps in which
the sampling interval is large. It may also be desired to normalise
the weights and combine this approach with an exponential factor as
in Section \ref{subsec:Online-trackers}.

\subsubsection{Other choices}

In some applications, we may want to put more emphasis on certain
time steps in a different way from the ones discussed above. For example,
in a surveillance situation, there may have been an incident at a
certain time step, around which tracking accuracy is of high importance,
being less relevant at other time steps.

\subsection{Metric decomposition\label{subsec:Metric-decomposition}}

Following \cite[Eq. (24)]{Angel20_d}, the time-weighted multi-dimensional
assignment metric can be decomposed as

\begin{align}
 & d(\mathbf{X},\mathbf{Y})\nonumber \\
 & =\min_{\substack{W^{k}\in\mathcal{W}_{\mathbf{X},\mathbf{Y}}\\
k=1,\ldots,T
}
}\left(\sum_{k=1}^{T}w_{1}^{k}\mathrm{l}^{k}\left(\mathbf{X},\mathbf{Y},W^{k}\right)^{p}\right.\nonumber \\
 & \quad+\sum_{k=1}^{T}w_{1}^{k}\mathrm{m}^{k}\left(\mathbf{X},\mathbf{Y},W^{k}\right)^{p}+\sum_{k=1}^{T}w_{1}^{k}\mathrm{f}^{k}\left(\mathbf{X},\mathbf{Y},W^{k}\right)^{p}\nonumber \\
 & \left.\quad+\sum_{k=1}^{T-1}w_{2}^{k}\mathrm{s}^{k}\left(W^{k},W^{k+1}\right)^{p}\right)^{1/p}\label{eq:metric_decomposition}
\end{align}
where $\mathrm{l}^{k}\left(\cdot\right)$ is the localisation cost
for properly detected targets at time $k$, $\mathrm{m}^{k}\left(\cdot\right)$
is the missed target cost at time $k$, $\mathrm{f}^{k}\left(\cdot\right)$
is the false target cost at time $k$ and $\mathrm{s}^{k}\left(\cdot\right)$
is the switching cost at time $k$. The definition of these costs
is provided in \cite[Sec. IV.C]{Angel20_d} and is not repeated here.
The LP metric (\ref{eq:LP_metric}) has the same decomposition, but
with soft assignments. 

\subsection{Metric inequalities\label{subsec:Inequalities}}

It is direct to see that if $\gamma\rightarrow0$, the metrics (\ref{eq:multi-dimensional-metric})
and (\ref{eq:LP_metric}) tend to the ($p$-th root) of the weighted
sum of the GOSPA costs (to the $p$-th power) across time
\begin{align}
d^{0}\left(\mathbf{X},\mathbf{Y}\right) & =\left(\sum_{k=1}^{T}w_{1}^{k}d_{G}\left(\tau^{k}\left(\mathbf{X}\right),\tau^{k}\left(\mathbf{Y}\right)\right)^{p}\right)^{1/p}.
\end{align}
It is also direct to establish that $d^{0}\left(\cdot,\cdot\right)$
is a lower bound on the trajectory metrics (\ref{eq:multi-dimensional-metric})
and (\ref{eq:LP_metric}) (which require $\gamma>0$). We can compute
$d^{0}\left(\mathbf{X},\mathbf{Y}\right)$ by solving $T$ 2-D assignment
problems, one for each GOSPA cost. Solving these $T$ 2-D assignment
problems is considerably faster than solving the multi-dimensional
assignment problem or the LP problem in (\ref{eq:multi-dimensional-metric})
and (\ref{eq:LP_metric}). However, the distance $d^{0}\left(\mathbf{X},\mathbf{Y}\right)$
neglects track switching costs and is not a metric.

If $\gamma\rightarrow\infty$, both the multi-dimensional assignment
metric and the LP metric tend to the same metric, which we refer to
as $d^{\infty}\left(\mathbf{X},\mathbf{Y}\right)$. The metric $d^{\infty}\left(\mathbf{X},\mathbf{Y}\right)$
can be computed with a single 2-D assignment problem \cite[Sec. IV.D]{Angel20_d},
which is fast to compute, but it does not allow trajectories to change
assignments across time, which is usually undesirable, see Figure
\ref{fig:OSPA-track-switching}.  Then, for fixed values of $c$,
$p$ and $\gamma$, the following inequalities hold
\begin{equation}
d^{0}\left(\mathbf{X},\mathbf{Y}\right)<\overline{d}\left(\mathbf{X},\mathbf{Y}\right)\leq d\left(\mathbf{X},\mathbf{Y}\right)\leq d^{\infty}\left(\mathbf{X},\mathbf{Y}\right).
\end{equation}

\section{Metrics on RFSs: MOT evaluation across multiple scenarios\label{sec:Metrics-on-RFS}}

The time-weighted trajectory metrics can be directly extended to metrics
on RFSs of trajectories by considering expected values \cite{Angel20_d}.
To be specific, Lemma 3 in \cite{Angel20_d} is also valid for the
time-weighted metrics, as well as for the OSPA$^{(2)}$ metric, and
it is reproduced next.
\begin{lem}
\label{lem:metrics_RFS}Given $1\leq p'<\infty$, $\left(\mathbb{E}\big[d(\mathbf{X},\mathbf{Y})^{p'}\big]\right)^{1/p'}$
and $\left(\mathbb{E}\big[\overline{d}(\mathbf{X},\mathbf{Y})^{p'}\big]\right)^{1/p'}$
are metrics on the space of random finite sets of trajectories with
finite moment $\mathbb{E}\left[\left|\cdot\right|^{p'/p}\right]$,
with $\mathbb{E}\left[\cdot\right]$ denoting expected value.
\end{lem}
The metrics on RFSs are important to evaluate algorithms when we consider
a fixed ground truth and we use Monte Carlo simulation to draw different
realisations of the measurements. In this case, each algorithm returns
a different estimate at each Monte Carlo run. Metrics on RFSs are
also important when we have a dataset, either simulated or obtained
via experiments, with several scenarios, each with a different ground
truth \cite{Caesar20}. For each scenario, each algorithm provides
an estimate, and we look for a unified measure of performance across
the different scenarios.

Consider the task of comparing the average performance of $J$ MOT
algorithms, and suppose we have a data set with $N$ scenarios with
ground truth sets of trajectories $\mathbf{X}_{1},...,\mathbf{X}_{N}$,
for which the $j$-th algorithm estimates $\mathbf{Y}_{1}^{j},...,\mathbf{Y}_{N}^{j}$.
We can interpret that $\mathbf{X}_{1},...,\mathbf{X}_{N}$ and $\mathbf{Y}_{1}^{j},...,\mathbf{Y}_{N}^{j}$
approximate the joint distribution $p\left(\mathbf{X},\mathbf{Y}^{j}\right)$
on the ground truth and the estimate of the $j$-th algorithm such
that
\begin{align}
p\left(\mathbf{X},\mathbf{Y}^{j}\right) & \approx\frac{1}{N}\sum_{i=1}^{N}\delta_{\left(\mathbf{X}_{i},\mathbf{Y}_{i}^{j}\right)}\left(\mathbf{X},\mathbf{Y}^{j}\right).
\end{align}
Then, the distance between RFSs $\mathbf{X}$ and $\mathbf{Y}^{j}$
becomes
\begin{align}
\left(\mathbb{E}\big[d(\mathbf{X},\mathbf{Y}^{j})^{p'}\big]\right)^{1/p'} & \approx\left(\frac{1}{N}\sum_{i=1}^{N}\big[d(\mathbf{X}_{i},\mathbf{Y}_{i}^{j})^{p'}\big]\right)^{1/p'}.\label{eq:Error_multiple_ground_truths}
\end{align}
Equation (\ref{eq:Error_multiple_ground_truths}) provides a metric-based
ranking of performance for the $J$ algorithms across the $N$ scenarios.

\section{Example\label{sec:Examples}}

\begin{figure}
\begin{centering}
\includegraphics[scale=0.3]{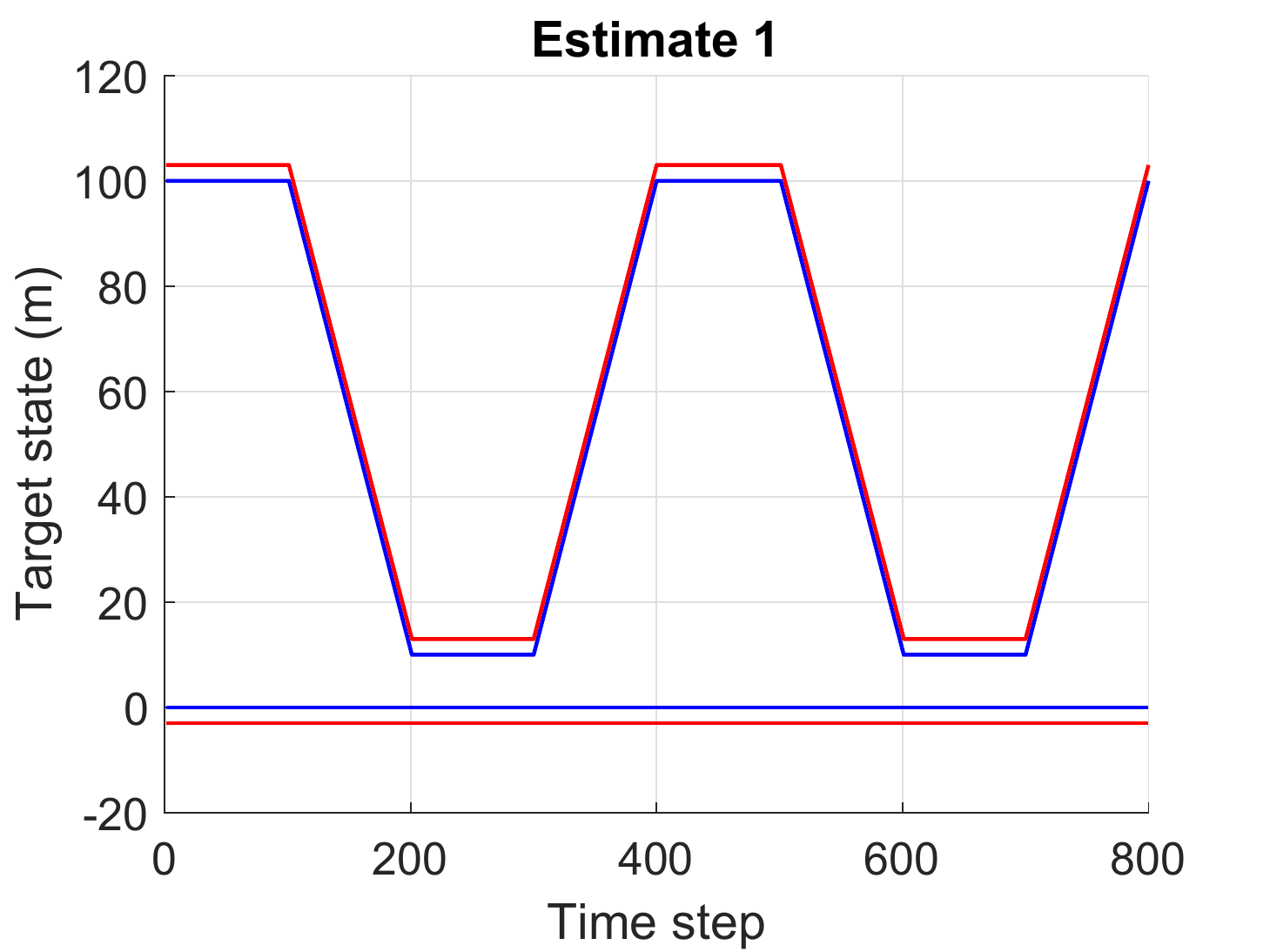}\includegraphics[scale=0.3]{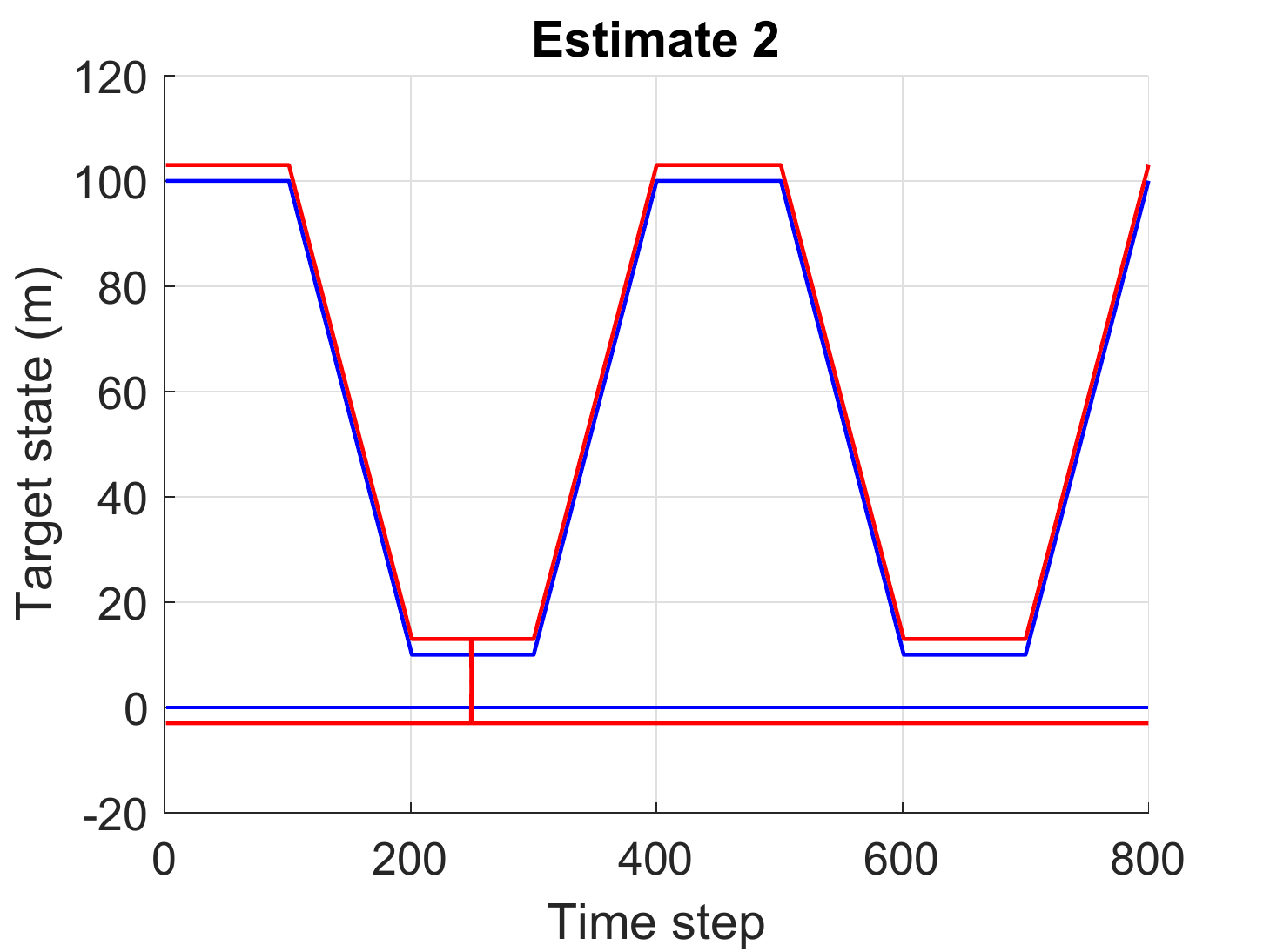}
\par\end{centering}
\begin{centering}
\includegraphics[scale=0.3]{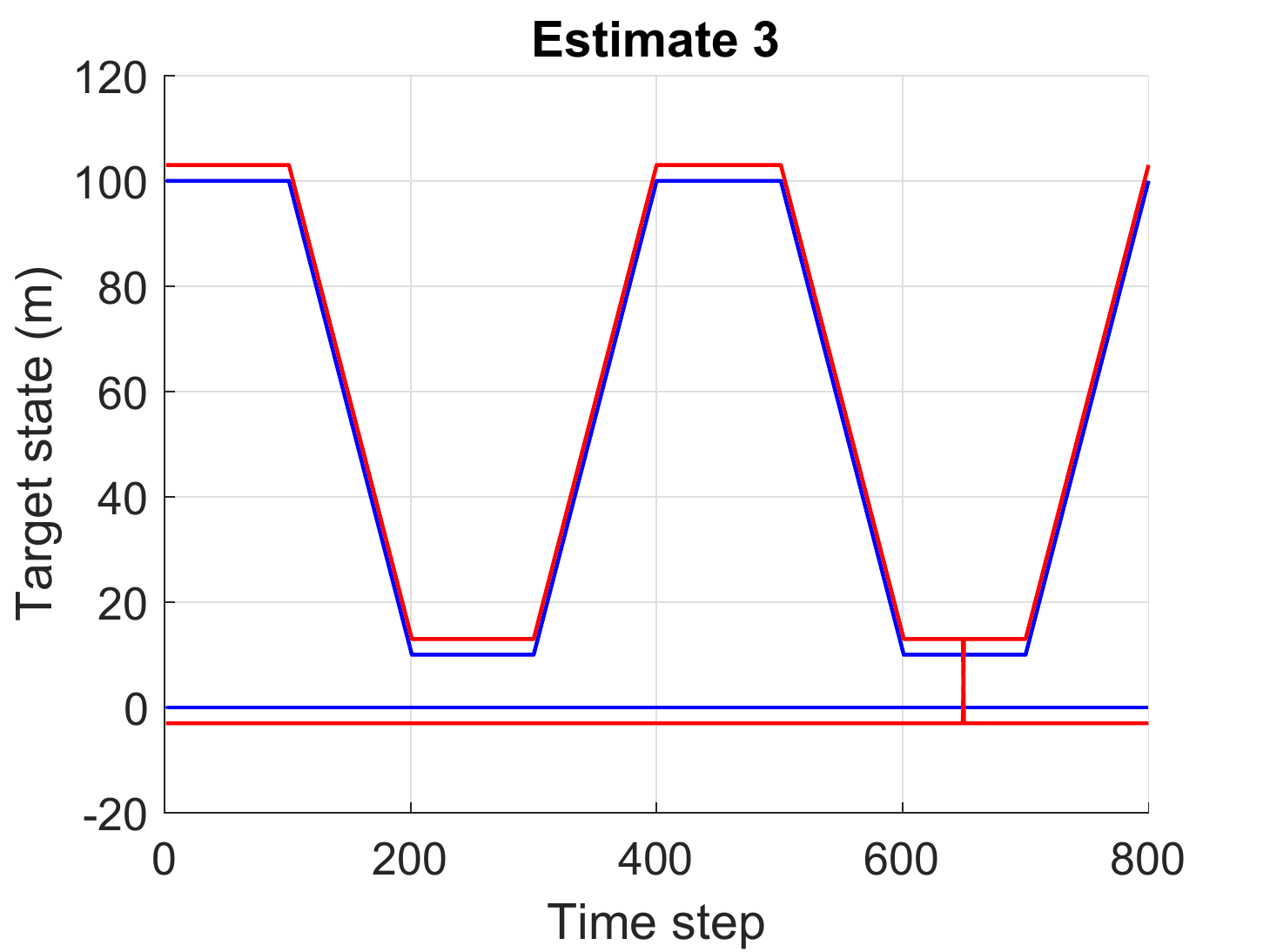}\includegraphics[scale=0.3]{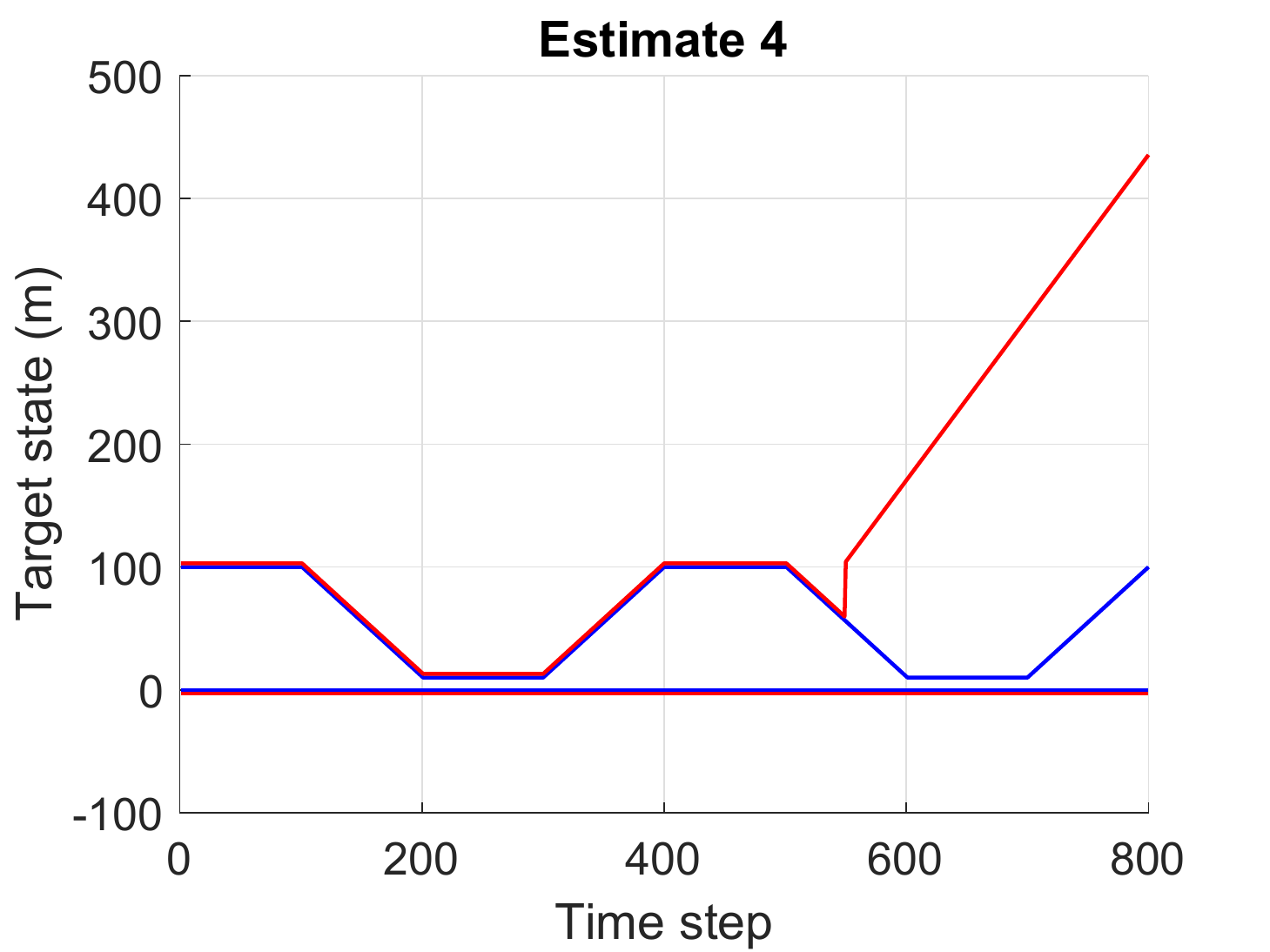}
\par\end{centering}
\caption{\label{fig:Scenario-four_estimates}Scenario of the example. We have
a ground truth set of trajectories with two trajectories (blue), and
four estimated sets of trajectories (red), each with two trajectories.
There is track switching in Estimates 2 and 3.}
\end{figure}

We proceed to illustrate how the trajectory metric (TM), the time-weighted
trajectory metric (TW-TM)\footnote{Matlab code for the time weighted LP trajectory metric is available
at https://github.com/Agarciafernandez/MTT.} and OSPA$^{(2)}$ differ in one example. We consider a single ground
truth with two one-dimensional trajectories, $T=800$ steps, and 4
estimates, each with two trajectories, as shown in Figure \ref{fig:Scenario-four_estimates}.
Estimate 1 (E1) estimates each trajectory with a deviation of 3 meters
at each time step. Estimate 2 (E2) and Estimate 3 (E3) are like E1
but with track switching at time step 250 and 650, respectively. Estimate
4 (E4) is like E1 but, from time step 550, one of the trajectories
is not estimated properly. 

Assuming that 3 meters is an acceptable localisation error, E1 would
be considered the best estimate and E4 the worst one in most MOT applications.
If we are assessing an online tracking algorithm and we want to penalise
errors at past time steps less, E2 should be better than E3. 

We use the TM with the Euclidean metric as the base distance and parameters
$c=5$, $p=1$, $\gamma=10$, and we normalise its result by the length
of the time window $T$. The TW-TM has the same parameters as the
TM with the normalised weights for online trackers in (\ref{eq:weight1_online_tracker_norm})
with $\rho=0.995$. The OSPA$^{(2)}$ metric also uses the Euclidean
metric, $c=5$, $p=1$ and is applied on the same time interval. We
have also tested the trajectory metrics with $\gamma=10^{8}$ to show
the results when assignments are not allowed to change in time, as
in OSPA$^{(2)}$. 

The resulting errors and the decomposition into (time-weighted) localisation
cost for properly detected targets (Loc.), cost for missed targets
(Mis.), cost for false targets (Fal.) and cost for track switches
(Swi.)  are shown in Table \ref{tab:Errors-and-their-decomposition},
and the resulting rankings in Table \ref{tab:Ranking_estimates}.
The TW-TM ranks the algorithms according to what we would expect for
an online tracker. The TM ranks the algorithm well, but indicates
that E2 and E3 have the same error, as for this metric, it does not
matter when the track switches take place. The error decomposition
provides relevant information about the estimates. On the contrary,
OSPA$^{(2)}$ does not rank the algorithms properly; E4 is ranked
as the second best algorithm despite the major errors in the estimation
of one trajectory. This type of localisation error is typically penalised
more than a track switch in most MOT applications \cite[Sec. 13.6]{Blackman_book99}.
Setting $\gamma=\infty$ or $\gamma$ sufficiently high in the trajectory
metrics also has undesirable effects in the ranking, due to the lack
of change in the assignments \cite{Blackman_book99}.

\begin{table*}
\caption{\label{tab:Errors-and-their-decomposition}Errors and their decomposition
for the estimates in Figure \ref{fig:Scenario-four_estimates}}

\centering{}%
\begin{tabular}{c|ccccc|ccccc||c|c|c}
\hline 
Metric &
\multicolumn{5}{c|}{TW-TM} &
\multicolumn{5}{c||}{TM} &
OSPA$^{(2)}$ &
TW-TM ($\gamma=10^{8}$) &
\multicolumn{1}{c}{TM ($\gamma=10^{8}$)}\tabularnewline
\hline 
 &
Tot. &
Loc. &
Mis. &
Fal. &
Swi. &
Tot. &
Loc. &
Mis. &
Fal. &
Swi. &
Tot. &
Tot. &
Tot.\tabularnewline
\hline 
$\mathrm{E}1$ &
6 &
6 &
0 &
0 &
0 &
6 &
6 &
0 &
0 &
0 &
3 &
6 &
6\tabularnewline
$\mathrm{E}2$ &
6.01 &
6.00 &
0 &
0 &
0.01 &
6.03 &
6.00 &
0 &
0 &
0.03 &
3.62 &
6.18 &
7.25\tabularnewline
$\mathrm{E3}$ &
6.05 &
6.00 &
0 &
0 &
0.05 &
6.03 &
6.00 &
0 &
0 &
0.03 &
3.38 &
7.84 &
6.76\tabularnewline
$\mathrm{E}4$ &
7.46 &
3.81 &
1.82 &
1.82 &
0 &
6.63 &
5.06 &
0.78 &
0.78 &
0 &
3.31 &
7.46 &
6.63\tabularnewline
\hline 
\end{tabular}
\end{table*}

\begin{table}
\caption{\label{tab:Ranking_estimates}Ranking of the estimates for the different
metrics}

\centering{}%
\begin{tabular}{cc}
\hline 
Metric &
Ranking\tabularnewline
\hline 
TW-TM &
E1-E2-E3-E4\tabularnewline
\multirow{2}{*}{TM} &
E1-E2-E3-E4\tabularnewline
 & E1-E3-E2-E4\tabularnewline
\hline 
\hline 
OSPA$^{(2)}$ &
E1-E4-E3-E2\tabularnewline
TW-TM ($\gamma=10^{8}$) &
E1-E2-E4-E3\tabularnewline
TM ($\gamma=10^{8}$) &
E1-E4-E3-E2\tabularnewline
\hline 
\end{tabular}
\end{table}

\begin{figure}
\begin{centering}
\includegraphics[scale=0.3]{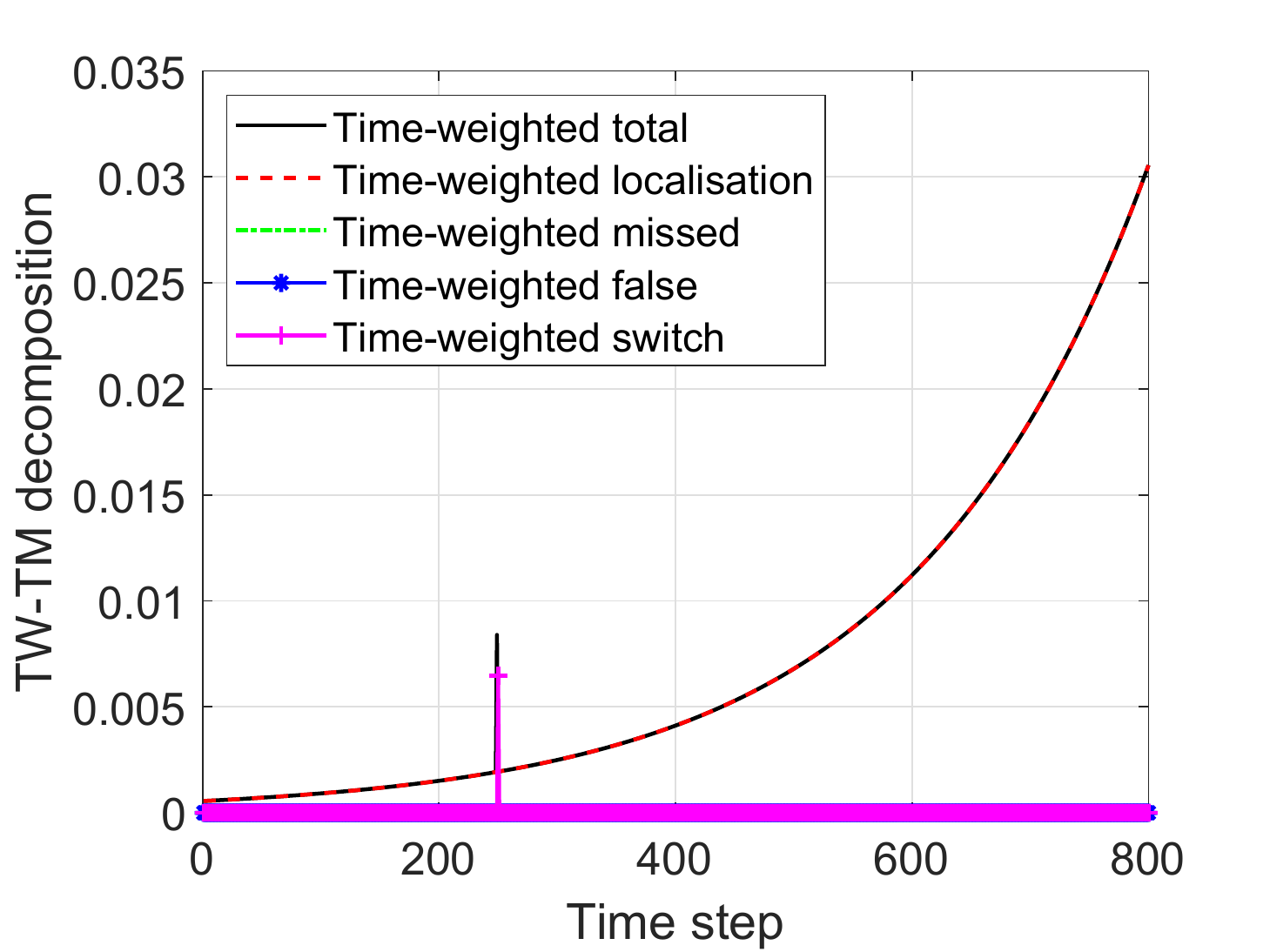}\includegraphics[scale=0.3]{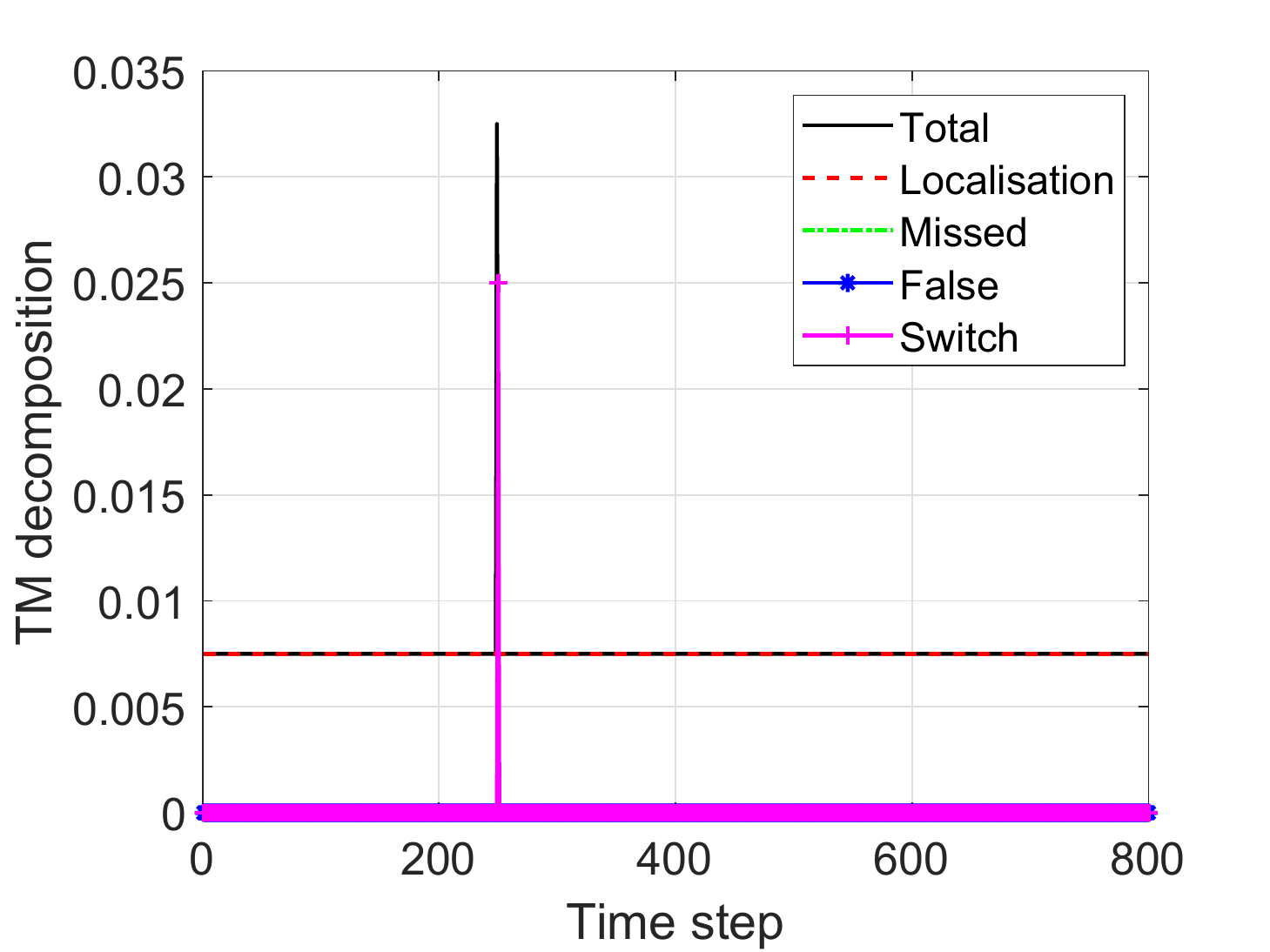}
\par\end{centering}
\caption{\label{fig:Decomposition-time}Decomposition of the error across time
for the time-weighted trajectory metric (left) and the trajectory
metric (right) for Estimate 2 in Figure \ref{fig:Scenario-four_estimates}.
In the TW-TM, the (time-weighted) costs increase with time due to
the considered weighting, see (\ref{eq:weight1_online_tracker_norm}). }

\end{figure}

Finally, we show the decomposition of the error across time, see Section
\ref{subsec:Metric-decomposition}, for the TW-TM and the TM in Figure
\ref{fig:Decomposition-time}. This decomposition provides useful
information, for example, we can locate when track switching occurs,
and determine that there are not any missed or false targets. 

\section{Conclusions\label{sec:Conclusions}}

This paper has extended the metric on sets of trajectories in \cite{Angel20_d}
to include weights to penalise the costs (localisation error, number
of missed/false targets, and track switches) at different time steps
unevenly. This extension adds flexibility to the trajectory metric
\cite{Angel20_d} to suit more user preferences to compute the error
of multiple object tracking algorithms. 

The metric can be computed solving a multi-dimensional assignment
problem, e.g., using the Viterbi algorithm. The LP relaxation of the
metric is also a metric and is computable in polynomial time.

\appendices{}

\section{\label{sec:AppendixA}}

In this appendix, we prove Proposition \ref{prop:LP_metric}. The
non-negativity, identity and symmetry properties of the metric in
(\ref{eq:LP_metric}) are straightforward. The proof of the triangle
inequality is analogous to the proof of the triangle inequality of
the trajectory metric \cite[App. B]{Angel20_d}, with some minor adaptations. 

The proof in this section is done for the LP metric, where the optimization
is over $W^{k}\in\mathcal{\overline{W}}_{\mathbf{X},\mathbf{Y}}$.
The proof is similar for the multi-dimensional assignment metric in
(\ref{eq:multi-dimensional-metric}), where the optimisation is over
$W^{k}\in\mathcal{W}_{\mathbf{X},\mathbf{Y}}$. We use $\overline{d}\left(\mathbf{X},\mathbf{Y},W^{1:T}\right)$
to denote the objective function in (\ref{eq:LP_metric}) as a function
of $W^{1:T}=\left(W^{1},...,W^{T}\right)$.

The outline of the triangle inequality proof is as follows. We assume
that we have three sets of trajectories $\mathbf{X}$, $\mathbf{Y}$,
$\mathbf{Z}$. Let $W_{\mathbf{X},\mathbf{Z}}^{\star,k}$ and $W_{\mathbf{Z},\mathbf{Y}}^{\star,k}$
be the weight matrices that minimise $\overline{d}\left(\mathbf{Y},\mathbf{Z},W^{1:T}\right)$
and $\overline{d}\left(\mathbf{Z},\mathbf{Y},W^{1:T}\right)$. Based
on $W_{\mathbf{X},\mathbf{Z}}^{\star,k}$ and $W_{\mathbf{Z},\mathbf{Y}}^{\star,k}$,
we construct a matrix $W_{\mathbf{X},\mathbf{Y}}^{k}\in\mathcal{\overline{W}}_{\mathbf{X},\mathbf{Y}}$
such that
\begin{align}
W_{\mathbf{X},\mathbf{Y}}^{k}(i,j) & =\sum_{l=1}^{n_{\mathbf{Z}}}W_{\mathbf{X},\mathbf{Z}}^{\star,k}(i,l)W_{\mathbf{Z},\mathbf{Y}}^{\star,k}(l,j)\label{eq:W_XY_append1}
\end{align}
for $i\in\{1,...,n_{\mathbf{X}}\}$ and $j\in\{1,...,n_{\mathbf{Y}}\}$,
and the rest of the elements of $W_{\mathbf{X},\mathbf{Y}}^{k}$ are
\begin{align}
 & W_{\mathbf{X},\mathbf{Y}}^{k}(i,j)\nonumber \\
 & =\begin{cases}
1-\sum_{j=1}^{n_{\mathbf{Y}}}W_{\mathbf{X},\mathbf{Y}}^{k}(i,j) & i\in\{1,...,n_{\mathbf{X}}\},j=n_{\mathbf{Y}}+1\\
1-\sum_{i=1}^{n_{\mathbf{X}}}W_{\mathbf{X},\mathbf{Y}}^{k}(i,j) & i=n_{\mathbf{X}}+1,j\in\{1,...,n_{\mathbf{Y}}\}\\
0 & i=n_{\mathbf{X}}+1,j=n_{\mathbf{Y}}+1.
\end{cases}\label{eq:W_XY_append2}
\end{align}
Then, we show that
\begin{align}
\overline{d}\left(\mathbf{X},\mathbf{Y},W_{\mathbf{X},\mathbf{Y}}^{1:T}\right) & \leq\overline{d}\left(\mathbf{X},\mathbf{Z}\right)+\overline{d}\left(\mathbf{Z},\mathbf{Y}\right).\label{eq:triIneq1}
\end{align}
By definition, $\overline{d}\left(\mathbf{X},\mathbf{Y}\right)\leq\overline{d}\left(\mathbf{X},\mathbf{Y},W_{\mathbf{X},\mathbf{Y}}^{1:T}\right)$,
which yields
\begin{align}
\overline{d}\left(\mathbf{X},\mathbf{Y}\right) & \leq\overline{d}\left(\mathbf{X},\mathbf{Z}\right)+\overline{d}\left(\mathbf{Z},\mathbf{Y}\right)
\end{align}
and finishes the proof of the triangle inequality. 

To prove (\ref{eq:triIneq1}), we show that for any $W_{\mathbf{X},\mathbf{Z}}^{k}\in\mathcal{\overline{W}}_{\mathbf{X},\mathbf{Z}}$
and $W_{\mathbf{Z},\mathbf{Y}}^{k}\in\mathcal{\overline{W}}_{\mathbf{Z},\mathbf{Y}}$,
and $W_{\mathbf{X},\mathbf{Y}}^{k}$ obtained using (\ref{eq:W_XY_append1})
and (\ref{eq:W_XY_append2}). we have
\begin{align}
\overline{d}\left(\mathbf{X},\mathbf{Y},W_{\mathbf{X},\mathbf{Y}}^{1:T}\right) & \leq\overline{d}\left(\mathbf{X},\mathbf{Z},W_{\mathbf{X},\mathbf{Z}}^{1:T}\right)+\overline{d}\left(\mathbf{Z},\mathbf{Y},W_{\mathbf{Z},\mathbf{Y}}^{1:T}\right).\label{eq:triIneq}
\end{align}
We prove (\ref{eq:triIneq}) in Section \ref{subsec:Proof-triangle-inequality}.
Before this, we provide three preliminary inequalities in Section
\ref{subsec:preliminary-ineq}.

\subsection{Preliminary inequalities\label{subsec:preliminary-ineq}}

The following switching cost inequality holds \cite[App. B]{Angel20_d}
\begin{align}
 & \sum_{i=1}^{n_{\mathbf{X}}}\sum_{j=1}^{n_{\mathbf{Y}}}\left|W_{\mathbf{X},\mathbf{Y}}^{k}\left(i,j\right)-W_{\mathbf{X},\mathbf{Y}}^{k+1}\left(i,j\right)\right|\nonumber \\
 & \leq\sum_{i=1}^{n_{\mathbf{X}}}\sum_{l=1}^{n_{\mathbf{Z}}}\left|W_{\mathbf{X},\mathbf{Z}}^{k}\left(i,l\right)-W_{\mathbf{X},\mathbf{Z}}^{k+1}\left(i,l\right)\right|\nonumber \\
 & \quad+\sum_{j=1}^{n_{\mathbf{Y}}}\sum_{l=1}^{n_{\mathbf{Z}}}\left|W_{\mathbf{Z},\mathbf{Y}}^{k}\left(l,j\right)-W_{\mathbf{Z},\mathbf{Y}}^{k+1}\left(l,j\right)\right|.\label{eq:switching_inequality}
\end{align}
The following inequality also holds \cite[App. B]{Angel20_d},
\begin{align}
 & \mathrm{tr}\big[\big(D_{\mathbf{X},\mathbf{Y}}^{k}\big)^{\dagger}W_{\mathbf{X},\mathbf{Y}}^{k}\big]\nonumber \\
 & \leq\sum_{i=1}^{n_{\mathbf{X}}}\sum_{j=1}^{n_{\mathbf{Y}}}\sum_{l=1}^{n_{\mathbf{Z}}}\left(d_{\mathbf{X},\mathbf{Z}}^{k}\left(i,l\right)+d_{\mathbf{Z},\mathbf{Y}}^{k}\left(l,j\right)\right)^{p}W_{\mathbf{X},\mathbf{Z}}^{k}\left(i,l\right)W_{\mathbf{Z},\mathbf{Y}}^{k}\left(l,j\right)\nonumber \\
 & \quad+\sum_{j=1}^{n_{\mathbf{Y}}}D_{\mathbf{Z},\mathbf{Y}}^{k}\left(n_{\mathbf{Z}}+1,j\right)W_{\mathbf{Z},\mathbf{Y}}^{k}\left(n_{\mathbf{Z}}+1,j\right)\nonumber \\
 & \quad+\sum_{j=1}^{n_{\mathbf{Y}}}\sum_{l=1}^{n_{\mathbf{Z}}}\left(d_{\mathbf{X},\mathbf{Z}}^{k}\left(n_{\mathbf{X}}+1,l\right)+d_{\mathbf{Z},\mathbf{Y}}^{k}\left(l,j\right)\right)^{p}\nonumber \\
 & \quad\times W_{\mathbf{X},\mathbf{Z}}^{k}\left(n_{\mathbf{X}}+1,l\right)W_{\mathbf{Z},\mathbf{Y}}^{k}\left(l,j\right)\nonumber \\
 & \quad+\sum_{i=1}^{n_{\mathbf{X}}}D_{\mathbf{X},\mathbf{Z}}^{k}\left(i,n_{\mathbf{Z}}+1\right)W_{\mathbf{X},\mathbf{Z}}^{k}\left(i,n_{\mathbf{Z}}+1\right)\nonumber \\
 & \quad+\sum_{i=1}^{n_{\mathbf{X}}}\sum_{l=1}^{n_{\mathbf{Z}}}\left(d_{\mathbf{X},\mathbf{Z}}^{k}\left(i,l\right)+d_{\mathbf{Z},\mathbf{Y}}^{k}\left(l,n_{\mathbf{Y}}+1\right)\right)^{p}\nonumber \\
 & \quad\times W_{\mathbf{X},\mathbf{Z}}^{k}\left(i,l\right)W_{\mathbf{Z},\mathbf{Y}}^{k}\left(l,n_{\mathbf{Y}}+1\right)\label{eq:localisation_inequality}
\end{align}
where $\ensuremath{d_{\mathbf{X},\mathbf{Y}}^{k}\left(i,j\right)=D_{\mathbf{X},\mathbf{Y}}^{k}(i,j)^{1/p}}$.

Finally, for $p\geq1$, and $a_{m},b_{m}\geq0$, the Minkowski inequality
\cite[pp. 165]{Kubrusly_book11} is
\begin{align}
\left(\sum_{m}\left[a_{m}+b_{m}\right]^{p}\right)^{1/p} & \leq\left(\sum_{m}a_{m}^{p}\right)^{1/p}+\left(\sum_{m}b_{m}^{p}\right)^{1/p}.\label{eq:Minkowski}
\end{align}

\subsection{Proof of (\ref{eq:triIneq})\label{subsec:Proof-triangle-inequality}}

From (\ref{eq:localisation_inequality}), we can write
\begin{align}
 & w_{1}^{k}\mathrm{tr}\big[\big(D_{\mathbf{X},\mathbf{Y}}^{k}\big)^{\dagger}W_{\mathbf{X},\mathbf{Y}}^{k}\big]\nonumber \\
 & \leq\sum_{i=1}^{n_{\mathbf{X}}}\sum_{j=1}^{n_{\mathbf{Y}}}\sum_{l=1}^{n_{\mathbf{Z}}}\left(a_{i,l,j}^{k}+b_{i,l,j}^{k}\right)^{p}+\sum_{j=1}^{n_{\mathbf{Y}}}\left(0+b_{j}^{k}\right)^{p}\nonumber \\
 & +\sum_{j=1}^{n_{\mathbf{Y}}}\sum_{l=1}^{n_{\mathbf{Z}}}\left(a_{l,j}^{k}+b_{l,j}^{k}\right)^{p}+\sum_{i=1}^{n_{\mathbf{X}}}\left(a_{i}^{k}+0\right)^{p}\nonumber \\
 & +\sum_{i=1}^{n_{\mathbf{X}}}\sum_{l=1}^{n_{\mathbf{Z}}}\left(a_{i,l}^{k}+b_{i,l}^{k}\right)^{p}\label{eq:localisation_inequality_ab}
\end{align}
where we have multiplied the cost by $w_{1}^{k}$ and it is direct
to identify the terms $a_{i,l,j}^{k}$ and $b_{i,l,j}^{k}$ once $w_{1}^{k}$
and the weight matrices are included inside the parenthesis for each
summation in (\ref{eq:localisation_inequality}).

Combining (\ref{eq:switching_inequality}) and (\ref{eq:localisation_inequality})
with $\overline{d}\left(\mathbf{X},\mathbf{Y},W_{\mathbf{X},\mathbf{Y}}^{1:T}\right)$,
we obtain
\begin{align}
 & \overline{d}\left(\mathbf{X},\mathbf{Y},W_{\mathbf{X},\mathbf{Y}}^{1:T}\right)\nonumber \\
 & \leq\left[\sum_{k=1}^{T}\left(\sum_{i=1}^{n_{\mathbf{X}}}\sum_{j=1}^{n_{\mathbf{Y}}}\sum_{l=1}^{n_{\mathbf{Z}}}\left(a_{i,l,j}^{k}+b_{i,l,j}^{k}\right)^{p}+\sum_{j=1}^{n_{\mathbf{Y}}}\left(0+b_{j}^{k}\right)^{p}\right.\right.\nonumber \\
 & +\sum_{j=1}^{n_{\mathbf{Y}}}\sum_{l=1}^{n_{\mathbf{Z}}}\left(a_{l,j}^{k}+b_{l,j}^{k}\right)^{p}+\sum_{i=1}^{n_{\mathbf{X}}}\left(a_{i}^{k}+0\right)^{p}\nonumber \\
 & \left.+\sum_{i=1}^{n_{\mathbf{X}}}\sum_{l=1}^{n_{\mathbf{Z}}}\left(a_{i,l}^{k}+b_{i,l}^{k}\right)^{p}\right)\nonumber \\
 & \quad+\sum_{k=1}^{T-1}\sum_{i=1}^{n_{\mathbf{X}}}\sum_{l=1}^{n_{\mathbf{Z}}}\underbrace{\frac{\gamma^{p}}{2}w_{2}^{k}\left|W_{\mathbf{X},\mathbf{Z}}^{k}\left(i,l\right)-W_{\mathbf{X},\mathbf{Z}}^{k+1}\left(i,l\right)\right|}_{\left(a_{i,l}^{k,2}+0\right)^{p}}\nonumber \\
 & \left.\quad+\sum_{k=1}^{T-1}\sum_{j=1}^{n_{\mathbf{Y}}}\sum_{l=1}^{n_{\mathbf{Z}}}\underbrace{\frac{\gamma^{p}}{2}w_{2}^{k}\left|W_{\mathbf{Z},\mathbf{Y}}^{k}\left(l,j\right)-W_{\mathbf{Z},\mathbf{Y}}^{k+1}\left(l,j\right)\right|}_{\left(0+b_{l,j}^{k,2}\right)^{p}}\right]^{1/p}.
\end{align}
We make use of the Minkowski inequality (\ref{eq:Minkowski}) to obtain
\begin{align}
 & \overline{d}\left(\mathbf{X},\mathbf{Y},W_{\mathbf{X},\mathbf{Y}}^{1:T}\right)\nonumber \\
 & \leq\left[\sum_{k=1}^{T}\left(\sum_{i=1}^{n_{\mathbf{X}}}\sum_{j=1}^{n_{\mathbf{Y}}}\sum_{l=1}^{n_{\mathbf{Z}}}\left(a_{i,l,j}^{k}\right)^{p}+\sum_{j=1}^{n_{\mathbf{Y}}}\sum_{l=1}^{n_{\mathbf{Z}}}\left(a_{l,j}^{k}\right)^{p}\right.\right.\nonumber \\
 & \left.+\sum_{i=1}^{n_{\mathbf{X}}}\left(a_{i}^{k}+0\right)^{p}+\sum_{i=1}^{n_{\mathbf{X}}}\sum_{l=1}^{n_{\mathbf{Z}}}\left(a_{i,l}^{k}\right)^{p}\right)\nonumber \\
 & \left.+\sum_{k=1}^{T-1}\sum_{i=1}^{n_{\mathbf{X}}}\sum_{l=1}^{n_{\mathbf{Z}}}\frac{\gamma^{p}}{2}w_{2}^{k}\left|W_{\mathbf{X},\mathbf{Z}}^{k}\left(i,l\right)-W_{\mathbf{X},\mathbf{Z}}^{k+1}\left(i,l\right)\right|\right]^{1/p}\nonumber \\
 & +\left[\sum_{k=1}^{T}\left(\sum_{i=1}^{n_{\mathbf{X}}}\sum_{j=1}^{n_{\mathbf{Y}}}\sum_{l=1}^{n_{\mathbf{Z}}}\left(b_{i,l,j}^{k}\right)^{p}+\sum_{j=1}^{n_{\mathbf{Y}}}\left(0+b_{j}^{k}\right)^{p}\right.\right.\nonumber \\
 & \left.+\sum_{j=1}^{n_{\mathbf{Y}}}\sum_{l=1}^{n_{\mathbf{Z}}}\left(b_{l,j}^{k}\right)^{p}+\left.\sum_{i=1}^{n_{\mathbf{X}}}\sum_{l=1}^{n_{\mathbf{Z}}}\left(b_{i,l}^{k}\right)^{p}\right)\right)\nonumber \\
 & \left.+\sum_{k=1}^{T-1}\sum_{j=1}^{n_{\mathbf{Y}}}\sum_{l=1}^{n_{\mathbf{Z}}}\frac{\gamma^{p}}{2}w_{2}^{k}\left|W_{\mathbf{Z},\mathbf{Y}}^{k}\left(l,j\right)-W_{\mathbf{Z},\mathbf{Y}}^{k+1}\left(l,j\right)\right|\right]^{1/p}.\label{eq:triangle_inequality_step1}
\end{align}

By using (\ref{eq:localisation_inequality}) and (\ref{eq:localisation_inequality_ab}),
we proceed to write the terms that include a coefficient $a_{i,l,j}^{k}$
in terms of the distances. We have
\begin{align*}
 & \sum_{i=1}^{n_{\mathbf{X}}}\sum_{j=1}^{n_{\mathbf{Y}}}\sum_{l=1}^{n_{\mathbf{Z}}}\left(a_{i,l,j}^{k}\right)^{p}+\sum_{j=1}^{n_{\mathbf{Y}}}\sum_{l=1}^{n_{\mathbf{Z}}}\left(a_{l,j}^{k}\right)^{p}\\
 & +\sum_{i=1}^{n_{\mathbf{X}}}\left(a_{i}^{k}+0\right)^{p}+\sum_{i=1}^{n_{\mathbf{X}}}\sum_{l=1}^{n_{\mathbf{Z}}}\left(a_{i,l}^{k}\right)^{p}\\
 & =\sum_{i=1}^{n_{\mathbf{X}}}\sum_{j=1}^{n_{\mathbf{Y}}}\sum_{l=1}^{n_{\mathbf{Z}}}w_{1}^{k}d_{\mathbf{X},\mathbf{Z}}^{k}\left(i,l\right)^{p}W_{\mathbf{X},\mathbf{Z}}^{k}\left(i,l\right)W_{\mathbf{Z},\mathbf{Y}}^{k}\left(l,j\right)\\
 & +\sum_{j=1}^{n_{\mathbf{Y}}}\sum_{l=1}^{n_{\mathbf{Z}}}w_{1}^{k}d_{\mathbf{X},\mathbf{Z}}^{k}\left(n_{\mathbf{X}}+1,l\right)^{p}W_{\mathbf{X},\mathbf{Z}}^{k}\left(n_{\mathbf{X}}+1,l\right)W_{\mathbf{Z},\mathbf{Y}}^{k}\left(l,j\right)\\
 & +\sum_{i=1}^{n_{\mathbf{X}}}w_{1}^{k}D_{\mathbf{X},\mathbf{Z}}^{k}\left(i,n_{\mathbf{Z}}+1\right)W_{\mathbf{X},\mathbf{Z}}^{k}\left(i,n_{\mathbf{Z}}+1\right)\\
 & +\sum_{i=1}^{n_{\mathbf{X}}}\sum_{l=1}^{n_{\mathbf{Z}}}w_{1}^{k}d_{\mathbf{X},\mathbf{Z}}^{k}\left(i,l\right)^{p}W_{\mathbf{X},\mathbf{Z}}^{k}\left(i,l\right)W_{\mathbf{Z},\mathbf{Y}}^{k}\left(l,n_{\mathbf{Y}}+1\right)\\
 & =w_{1}^{k}\mathrm{tr}\big[\big(D_{\mathbf{X},\mathbf{Z}}^{k}\big)^{\dagger}W_{\mathbf{X},\mathbf{Z}}^{k}\big]
\end{align*}
where the last equality follows from \cite[Eq. (52)]{Angel20_d}.
Similarly, the sums of the terms with coefficient $b_{i,l,j}^{k}$
sum to $w_{1}^{k}\mathrm{tr}\big[\big(D_{\mathbf{Z},\mathbf{Y}}^{k}\big)^{\dagger}W_{\mathbf{Z},\mathbf{Y}}^{k}\big]$.
Substituting these values into (\ref{eq:triangle_inequality_step1})
yields (\ref{eq:triIneq}), which implies that the triangle inequality
holds.

\bibliographystyle{IEEEtran}
\bibliography{0C__Trabajo_laptop_Referencias_Referencias}

\begin{thebibliography}{10}
\providecommand{\url}[1]{#1}
\csname url@samestyle\endcsname
\providecommand{\newblock}{\relax}
\providecommand{\bibinfo}[2]{#2}
\providecommand{\BIBentrySTDinterwordspacing}{\spaceskip=0pt\relax}
\providecommand{\BIBentryALTinterwordstretchfactor}{4}
\providecommand{\BIBentryALTinterwordspacing}{\spaceskip=\fontdimen2\font plus
\BIBentryALTinterwordstretchfactor\fontdimen3\font minus
  \fontdimen4\font\relax}
\providecommand{\BIBforeignlanguage}[2]{{%
\expandafter\ifx\csname l@#1\endcsname\relax
\typeout{** WARNING: IEEEtran.bst: No hyphenation pattern has been}%
\typeout{** loaded for the language `#1'. Using the pattern for}%
\typeout{** the default language instead.}%
\else
\language=\csname l@#1\endcsname
\fi
#2}}
\providecommand{\BIBdecl}{\relax}
\BIBdecl

\bibitem{Angel20_d}
A.~F. Garc{\'\i}a-Fern{\'a}ndez, A.~S. Rahmathullah, and L.~Svensson, ``A
  metric on the space of finite sets of trajectories for evaluation of
  multi-target tracking algorithms,'' \emph{IEEE Transactions on Signal
  Processing}, vol.~68, pp. 3917--3928, 2020.

\bibitem{Mahler_book14}
R.~P.~S. Mahler, \emph{Advances in Statistical Multisource-Multitarget
  Information Fusion}.\hskip 1em plus 0.5em minus 0.4em\relax Artech House,
  2014.

\bibitem{Angel20_b}
A.~F. García-Fernández, L.~Svensson, and M.~R. Morelande, ``Multiple target
  tracking based on sets of trajectories,'' \emph{IEEE Transactions on
  Aerospace and Electronic Systems}, vol.~56, no.~3, pp. 1685--1707, Jun. 2020.

\bibitem{Milan16_arxiv}
\BIBentryALTinterwordspacing
A.~Milan, L.~Leal-Taixé, I.~Reid, S.~Roth, and K.~Schindler, ``{MOT16:} a
  benchmark for multi-object tracking.'' [Online]. Available:
  \url{https://arxiv.org/abs/1603.00831}
\BIBentrySTDinterwordspacing

\bibitem{Apostol_book74}
T.~M. Apostol, \emph{Mathematical Analysis}.\hskip 1em plus 0.5em minus
  0.4em\relax Addison Wesley, 1974.

\bibitem{Blackman_book99}
S.~Blackman and R.~Popoli, \emph{Design and Analysis of Modern Tracking
  Systems}.\hskip 1em plus 0.5em minus 0.4em\relax Artech House, 1999.

\bibitem{Fridling91}
B.~E. Fridling and O.~E. Drummond, ``Performance evaluation methods for
  multiple-target-tracking algorithms,'' vol. 1481, 1991, pp. 371--383.

\bibitem{Drummond92}
O.~E. Drummond and B.~E. Fridling, ``Ambiguities in evaluating performance of
  multiple target tracking algorithms,'' in \emph{Proceedings of the SPIE
  conference}, 1992, pp. 326--337.

\bibitem{Schuhmacher08_b}
D.~Schuhmacher and A.~Xia, ``A new metric between distributions of point
  processes,'' \emph{Advances in Applied Probability}, vol.~40, no.~3, pp.
  651--672, Sep. 2008.

\bibitem{Schuhmacher08}
D.~Schuhmacher, B.-T. Vo, and B.-N. Vo, ``A consistent metric for performance
  evaluation of multi-object filters,'' \emph{IEEE Transactions on Signal
  Processing}, vol.~56, no.~8, pp. 3447--3457, Aug. 2008.

\bibitem{Angel19_d}
A.~F. García-Fernández and L.~Svensson, ``Spooky effect in optimal {OSPA}
  estimation and how {GOSPA} solves it,'' in \emph{Proceedings on the 22nd
  International Conference on Information Fusion}, 2019.

\bibitem{Rahmathullah17}
A.~S. Rahmathullah, A.~F. García-Fernández, and L.~Svensson, ``Generalized
  optimal sub-pattern assignment metric,'' in \emph{20th International
  Conference on Information Fusion}, 2017, pp. 1--8.

\bibitem{Bernardin08}
K.~Bernardin and R.~Stiefelhagen, ``Evaluating multiple object tracking
  performance: {T}he {CLEAR MOT} metrics,'' \emph{EURASIP Journal on Image and
  Video Processing}, vol. 2008, pp. 1--10, 2008.

\bibitem{Ristani16}
E.~Ristani, F.~Solera, R.~Zou, R.~Cucchiara, and C.~Tomasi, ``Performance
  measures and a data set for multi-target, multi-camera tracking,'' in
  \emph{ECCV 2016 Workshops}, pp. 17--35.

\bibitem{Luiten20}
J.~Luiten \emph{et~al.}, ``{HOTA}: A higher order metric for evaluating
  multi-object tracking,'' \emph{International Journal of Computer Vision}, pp.
  1--31, 2020.

\bibitem{Ristic11}
B.~Ristic, B.-N. Vo, D.~Clark, and B.-T. Vo, ``A metric for performance
  evaluation of multi-target tracking algorithms,'' \emph{IEEE Transactions on
  Signal Processing}, vol.~59, no.~7, pp. 3452--3457, July 2011.

\bibitem{Canavan09}
R.~{Canavan}, C.~{McCullough}, and W.~J. {Farrell}, ``Track-centric metrics for
  track fusion systems,'' in \emph{12th International Conference on Information
  Fusion}, 2009, pp. 1147--1154.

\bibitem{Silbert09}
M.~{Silbert}, ``A robust method for computing truth-to-track assignments,'' in
  \emph{12th International Conference on Information Fusion}, 2009, pp.
  1658--1664.

\bibitem{Manson92}
K.~Manson and P.~O'Kane, ``Taxonomic performance evaluation for multitarget
  tracking systems,'' \emph{IEEE Transactions on Aerospace and Electronic
  Systems}, vol.~28, no.~3, pp. 775--787, July 1992.

\bibitem{Beard20}
M.~{Beard}, B.~T. {Vo}, and B.~{Vo}, ``A solution for large-scale multi-object
  tracking,'' \emph{IEEE Transactions on Signal Processing}, vol.~68, pp.
  2754--2769, 2020.

\bibitem{Bento_draft16}
\BIBentryALTinterwordspacing
J.~Bento and J.~J. Zhu, ``A metric for sets of trajectories that is practical
  and mathematically consistent,'' 2018. [Online]. Available:
  \url{https://arxiv.org/abs/1601.03094}
\BIBentrySTDinterwordspacing

\bibitem{Hero_book08}
A.~O. Hero~{III}, D.~A. Castañón, D.~Cochran, and K.~Kastella,
  \emph{Foundations and Applications of Sensor Management}.\hskip 1em plus
  0.5em minus 0.4em\relax Springer, 2008.

\bibitem{Granstrom18}
K.~Granstr{ö}m, L.~Svensson, Y.~Xia, J.~L. Williams, and A.~F.
  García-Fernández, ``Poisson multi-{B}ernoulli mixture trackers: continuity
  through random finite sets of trajectories,'' in \emph{21st International
  Conference on Information Fusion}, 2018, pp. 973--981.

\bibitem{Granstrom19_prov2}
\BIBentryALTinterwordspacing
K.~Granstr{ö}m, L.~Svensson, Y.~Xia, J.~Williams, and A.~F.
  Garc{í}a-Fern{á}ndez, ``Poisson multi-{B}ernoulli mixtures for sets of
  trajectories,'' 2019. [Online]. Available:
  \url{https://arxiv.org/abs/1912.08718}
\BIBentrySTDinterwordspacing

\bibitem{Tang19b}
Y.~C. Tang and R.~Salakhutdinov, ``Multiple futures prediction,'' in \emph{33rd
  Conference on Neural Information Processing Systems}, 2019.

\bibitem{Angel20_f}
A.~F. García-Fernández and S.~Maskell, ``Continuous-discrete trajectory {PHD}
  and {CPHD} filters,'' in \emph{23rd International Conference on Information
  Fusion}, 2020, pp. 1--8.

\bibitem{Caesar20}
H.~Caesar \emph{et~al.}, ``nu{S}cenes: A multimodal dataset for autonomous
  driving,'' in \emph{IEEE/CVF Conference on Computer Vision and Pattern
  Recognition (CVPR)}, 2020, pp. 11\,618--11\,628.

\bibitem{Kubrusly_book11}
C.~S. Kubrusly, \emph{The Elements of Operator Theory}.\hskip 1em plus 0.5em
  minus 0.4em\relax Springer Science + Business Media, 2011.

\end{thebibliography}

\end{document}